\documentclass{article} %
\usepackage{iclr2027_conference,times}

\usepackage{amsmath,amsfonts,bm}

\def\eqref#1{equation~\ref{#1}}

\def\1{\bm{1}}

\DeclareMathAlphabet{\mathsfit}{\encodingdefault}{\sfdefault}{m}{sl}
\SetMathAlphabet{\mathsfit}{bold}{\encodingdefault}{\sfdefault}{bx}{n}

\usepackage[linktoc=page]{hyperref}
\usepackage{url}
\usepackage{graphicx}
\usepackage{booktabs}
\usepackage[table]{xcolor}
\usepackage{array}
\usepackage{subcaption}
\usepackage{duckuments}
\usepackage{cleveref}
\usepackage{longtable}
\usepackage{tabularx}
\usepackage{etoc}
\usepackage{fontawesome5}
\usepackage{multirow}
\usepackage{float}
\floatstyle{plaintop}
\restylefloat{table}
\usepackage[tableposition=top]{caption}

\title{Removing Timing Shortcuts \\ Improves Non-Invasive Brain-to-Text}

\author{Dulhan Jayalath \& Oiwi Parker Jones \\
Neural Processing Lab (PNPL\includegraphics[height=2.2ex]{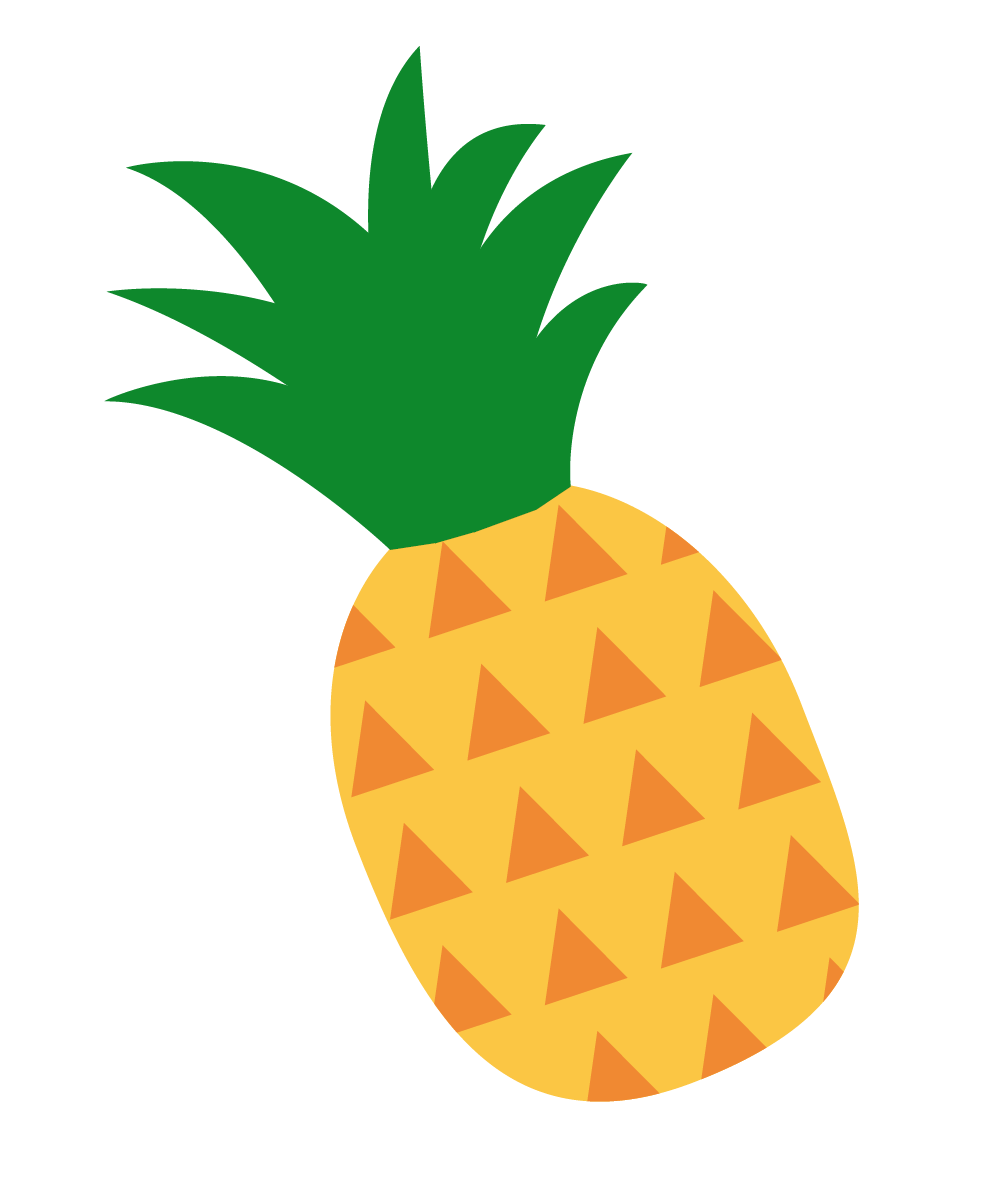}) \\ Department of Engineering Science, University of Oxford \\
{\texttt{\{dulhan, oiwi\}@robots.ox.ac.uk}}
}

\iclrfinalcopy %
\begin{document}

\begin{center}
\maketitle
\end{center}

\vspace{-1em}
\begin{abstract}

{
\looseness=-1 
We find that major reported improvements in decoding words from non-invasive brain recordings are largely reproducible without any brain data.
In the influential work of \citet{dAscoli2025TowardsDI}, time series of brain activity from subjects perceiving continuous speech are segmented into fixed-length windows starting at each word.
A neural network then generates predictions for all of the words in a sentence together.
Neighbouring windows partially overlap, implicitly revealing the interval between words.
Since these intervals indicate the duration of the words spoken, and different words tend to have different durations---for example, ``the'' is much shorter than ``supercalifragilisticexpialidocious''---the neural network can improve its predictions of words without relying on the underlying brain activity. Consistent with this, 
\textbf{the method reaches 22.0\% balanced accuracy on synthetic signals containing no brain information, compared with 22.3\% on real brain recordings}.
To prevent the network from learning this shortcut, we make a single, simple change. Instead of jointly encoding all windows in a sentence, we process each independently.
As a result, the neural network achieves better performance by learning underlying word-specific information from brain recordings.
This makes two existing strategies become much more effective than before.
Both aggregating predictions from distinct neural responses to the same word and using a pretrained LLM as a linguistic prior now substantially improve results.
On our clinically motivated perceived speech benchmark, this simple recipe (\emph{SimpleB2T}) achieves a \textbf{word error rate of 36.6\%} with five observations per word, approaching past invasive speech decoding performance, albeit under different conditions.
The results in this work expose an important shortcut in brain-to-text decoding and show that removing it leads to a simple and considerably more effective strategy.

\begin{center}
\textbf{Code \& Notebooks} \faGithub\, \href{https://github.com/neural-processing-lab/SimpleB2T}{\textcolor{black}{github.com/neural-processing-lab/SimpleB2T}} \\
\textbf{Benchmark} \faDatabase\, \href{https://github.com/neural-processing-lab/pnpl/blob/main/docs/clinical_communication.md}{github.com/neural-processing-lab/pnpl}
\end{center}
}

\end{abstract}

\section{Introduction}

\looseness=-1 Restoring communication to people who have lost the ability to speak by decoding brain activity into speech is a north star in brain--computer interface (BCI) research. Non-invasive approaches based on magnetoencephalography (MEG) or electroencephalography (EEG) offer a safe route
towards this goal, but must recover speech information from weak and spatially blurred signals measured outside the brain. Nevertheless, the state of the art in the field has progressed from detecting the presence of speech \citep{Dash2020NeuroVADRV}, to matching audio with corresponding neural responses~\citep{Defossez2022DecodingSP}, and more recently to decoding individual words \citep{dAscoli2025TowardsDI}. 

In this paper, we study \emph{word-aligned brain-to-text (B2T)} from perceived speech in non-invasive brain recordings. Here, a subject listens to speech (e.g. from an audiobook) or reads text while their brain activity is recorded with M/EEG and the task is to decode the words they perceived from their brain activity, knowing only when they perceived each word. Perceived speech is often used as a
stepping stone towards the grander goal of decoding internal speech, such as inner monologues, because it provides stronger and more
easily aligned neural responses \citep{Martin2014DecodingSF}. Thus, perceived speech is a common test-bed for developing non-invasive speech
decoding methods \citep{Defossez2022DecodingSP, Tang2022SemanticRO, ozdogan2025libribrain, mantegna2026pnplcompetition}.

\looseness=-1 One influential recent direction in word-aligned B2T, introduced by \citet{dAscoli2025TowardsDI}, has been to jointly decode neural responses to all words in a sentence from a continuous
M/EEG time series. In this setup, \citeauthor{dAscoli2025TowardsDI} extract a fixed-length window from the continuous recording at each word onset and then jointly encode all of the windows in a sentence with a neural network, predicting all of the words in the sentence at once. Predicting all of the words in the sentence together improves word classification by an average of 50\% compared with decoding each window independently \citep{dAscoli2025TowardsDI}. The setup has since been extended, evaluated, and
incorporated into a series of subsequent non-invasive brain-to-text studies and
benchmarks
\citep{zhang2025thought,jayalath2026meg,wang2026margin,li2026hdnd,
jayalath2025unlocking,ozdogan2025libribrain,
mantegna2026pnplcompetition,landau2026semantic,
banville2026neuralbench,Lvy2026NoninvasiveDO,mantegna2026,jayalath2026common}.
However, we identify an important property of this approach. Neighbouring
fixed-length windows typically overlap as the windows are longer than the individual words' durations. The same signal samples therefore appear at different relative positions in adjacent inputs, revealing the interval between word onsets and indicating the duration of words. Since typical word duration differs across words, the neural network can use this timing information to narrow the set of plausible words and improve its predictions, without using brain activity. This is an instance of \emph{shortcut learning} \citep{Geirhos2020ShortcutLI}, where a model can exploit an unintended decision rule that does not transfer to the intended use of the model.

\begin{figure}
    \centering
    \includegraphics[width=1.0\linewidth, trim=0 10 0 0, clip]{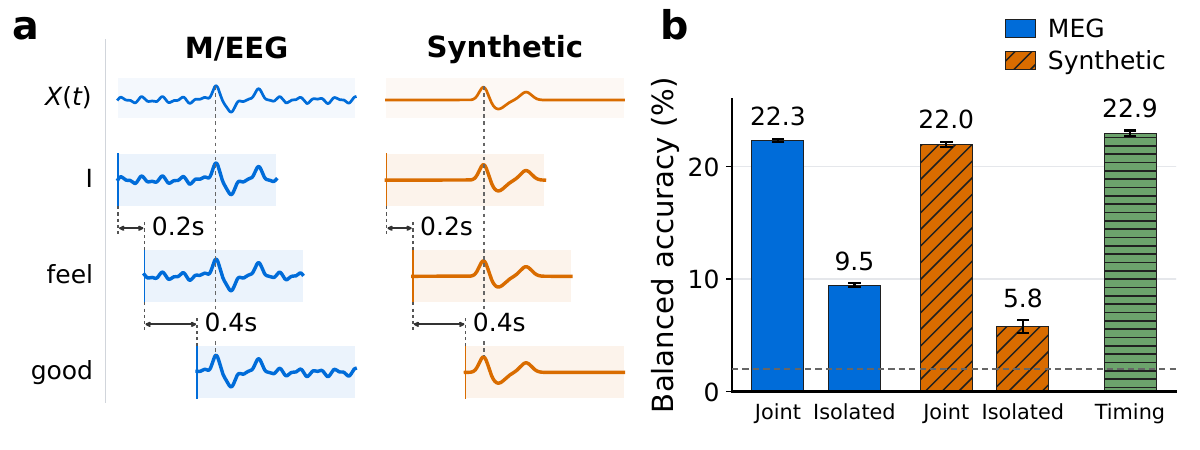}
\caption{\looseness=-1 \textbf{Overlapping word-aligned windows leak word duration.}
\textbf{(a)} Overlapping word-aligned windows contain shared samples whose relative offsets reveal the duration of words.
\textbf{(b)} To test whether jointly decoding words uses neural information, we compare MEG with synthetic signals having no relation to brain activity. With MEG, \emph{isolated} means we decode each window independently; with synthetic data, it means we make windows independent. \emph{Timing} tests what happens when the same model receives an encoding of the interval between words directly. Jointly decoding words reaches nearly the same accuracy on synthetic and real inputs, whereas removing the shortcut substantially reduces performance. Error bars indicate $\pm1 \sigma$ across training seeds.}
    \label{fig:controls}
    \vspace{-2em}
\end{figure}

We find that this effect is large enough to reproduce almost the entire gain
from jointly decoding words without using brain activity at all (Figure~\ref{fig:controls}). On real MEG,
jointly decoding words reaches 22.3\% balanced word accuracy, compared with 9.5\% when words are decoded independently. When we replace the MEG with a synthetic continuous signal
that contains no information about the stimulus but preserves the same window
overlap, the decoder reaches 22.0\%. Removing the overlap structure
reduces accuracy to 5.8\%. Thus, most of the apparent benefit of jointly decoding words can be recovered from the information exposed by overlapping inputs alone.

\looseness=-1 We explore the simplest solution to this problem: decode words individually instead of jointly so that the neural network can never learn this shortcut. We otherwise retain the setup of \citet{dAscoli2025TowardsDI}. Once words are decoded independently, two techniques that previously provided limited improvements become much more effective. 
Firstly, multiple occurrences of distinct neural responses to the same stimulus are commonly used in non-invasive BCIs to
improve signal-to-noise ratio \citep{Farwell1988TalkingOT}, and we apply the same principle here by
combining predictions from multiple observations of the same word.
Secondly, invasive speech BCIs routinely combine neural predictions with explicit linguistic priors by leveraging LLMs while non-invasive methods have so far struggled to do the same \citep{jayalath2025unlocking}. We find that predictions driven partly by timing are poorly suited to aggregation or combination with a linguistic prior, whereas when the shortcut is removed, predictions provide information that can be combined effectively with both. Compared with jointly decoding words, both strategies now yield much larger benefits from only a few aggregated observations. On a core subset of clinically motivated sentences, this approach, which we refer to as \emph{SimpleB2T}, achieves 65.6\% WER with
one observation per word and 36.6\% with five observations.

\looseness=-1 Our work makes three points: \textbf{(1)} major improvements in non-invasive word-aligned B2T are largely reproducible without brain information; \textbf{(2)} we establish controls that expose the source of this shortcut as overlapping word-aligned windows; and \textbf{(3)} decoding words independently provides a simple fix that removes the shortcut, forcing the model to use information from brain activity and making aggregation of predictions and combination with a linguistic prior much more effective.

\section{Word Duration Leakage in Word-Aligned Decoding}
\label{sec:leakage}

\looseness=-1 We revisit the setup of \citet{dAscoli2025TowardsDI}
and show how neighbouring, overlapping word-aligned windows reveal the duration of words to
the decoder. We then test whether this non-neural information is
sufficient to explain the large improvements from jointly decoding words.

\subsection{Decoding Words Jointly from Neural Responses}
\label{sec:contextual_decoding}

We describe the task and decoder of \citet{dAscoli2025TowardsDI}. In word-aligned B2T from perceived speech, the task is to decode the words a subject perceives through an auditory or visual stimulus from their recorded brain activity. The word-aligned nature of the task assumes that the onset time of each perceived word in the stimulus is known, but no further information is assumed.

Let $X(t)\in\mathbb{R}^{C}$ denote a continuous M/EEG recording with $C$ channels and
$t_i$ the known onset of word $w_i$. For each word, the model extracts a
three-second window beginning at its onset,
\begin{equation}
    x_i = X[t_i:t_i + 3\text{s}] \in \mathbb{R}^{C\times L},
\end{equation}
where $L$ is the number of time samples in three seconds. Windows $x_1,\dots,x_n$ from all words $w_1,\ldots,w_n$ in a sentence are then jointly encoded by a neural network into representations $\hat z_1,\dots, \hat z_n$ where these representations are semantic word embeddings.
For each word $w$, a target embedding $e_w$ is
obtained by encoding $w$ with T5-large \citep{raffel2020}. The neural network is trained with a contrastive objective which encourages the predicted representation
$\hat z_i$ to be similar to the target embedding $e_{w_i}$ and dissimilar to
embeddings of other words in the batch.

At test time, the model predicts a word by nearest-neighbour retrieval. Given a candidate
vocabulary $\mathcal{V}$, each predicted embedding is compared with the
corresponding frozen T5 embeddings using cosine similarity, and the
highest-scoring candidate is selected:
\begin{equation}
    \hat w_i
    =
    \operatorname*{arg\,max}_{w \in \mathcal{V}}
    \cos(\hat z_i, e_w).
    \label{eq:contextual_retrieval}
\end{equation}
\looseness=-1 Therefore, the neural network predicts a point in the T5 embedding space and retrieves the nearest candidate
word at evaluation. Jointly mapping all inputs $x_1,\dots,x_n$ for a sentence to $\hat z_1,\dots, \hat z_n$, instead of mapping each word-aligned window $x_i$ to $\hat z_i$ individually, yields a 50\% improvement on average in standard evaluations with continuous brain recordings \citep{dAscoli2025TowardsDI}.

\subsection{Overlapping Windows Reveal Word Duration}
An important property of this construction is that consecutive word windows
are not independent. Natural speech contains several words within three seconds, so adjacent windows
overlap:
\begin{equation}
    x_i = X[t_i:t_i+3\text{s}],
    \qquad
    x_{i+1} = X[t_{i+1}:t_{i+1}+3\text{s}],
\end{equation}
where $t_{i+1}$ is often much less than $t_i + 3\text{s}$. The same underlying samples therefore occur in neighbouring inputs at a relative
displacement of $t_{i+1}-t_i$.

Indeed, this interval can be recovered directly from two adjacent inputs. In
discrete time, let $d$ denote the number of samples between their onsets.
Because the windows are extracted from the same continuous recording,
$x_i[n+d]=x_{i+1}[n]$ throughout their overlapping region. Hence
\begin{equation}
    d =
    \arg\min_{\ell}
    \sum_c \sum_n
    \left(x^c_i[n+\ell] - x^c_{i+1}[n]\right)^2,
\end{equation}
where $c$ is the sensor channel, up to preprocessing and repeated signal patterns. Thus, recovering the interval between words (and thereby word duration) requires only matching the shared samples between neighbouring windows and is trivially solvable by minimising a sum of squared differences. 

When the input construction is applied to the natural speech data in this work, 99.9996\% of adjacent word
pairs have overlapping windows and these share 90.6\% of their samples on average. Moreover, the interval between word onsets correlates strongly with word duration ($r=0.90$; Appendix~\ref{app:timing_sanity}).

This raises a simple question: how much of the reported improvement from jointly decoding words comes from neural information, and how much can be recovered from the timing information exposed by overlapping windows?

\subsection{Synthetic Signals Reproduce Decoding Improvements}
\label{sec:timing_audit}

We test whether improvements from jointly decoding words require
neural information. The concern is that fixed-length windows for neighbouring
words overlap, so decoding words jointly may use word durations revealed by shared signal samples. We therefore construct controls that preserve or remove this overlap, allowing us to test both whether overlap alone can reproduce this gain and whether jointly decoding words remains useful when it is removed.

\paragraph{Experimental setup.}
We train variants of \citeauthor{dAscoli2025TowardsDI}'s decoder and evaluate them on subject 0 of LibriBrain100 \citep{mantegna2026}, the largest publicly available neural speech decoding dataset at the time of writing. In this dataset, the subject listened to around 80 hours of auditory stimuli from audiobooks, podcasts, and random spoken sentences while their brain activity was recorded with MEG. We split the dataset such that no stimuli overlap between splits and all examples from a session belong exclusively to training, validation, or test. We report balanced top-1 word accuracy over the fifty most frequent words in the dataset. Exact splits are in Appendix~\ref{app:data} and additional experimental details are in Appendix~\ref{app:expdetails}.

\begin{figure}[t]
    \centering
    \includegraphics[width=1.0\linewidth]{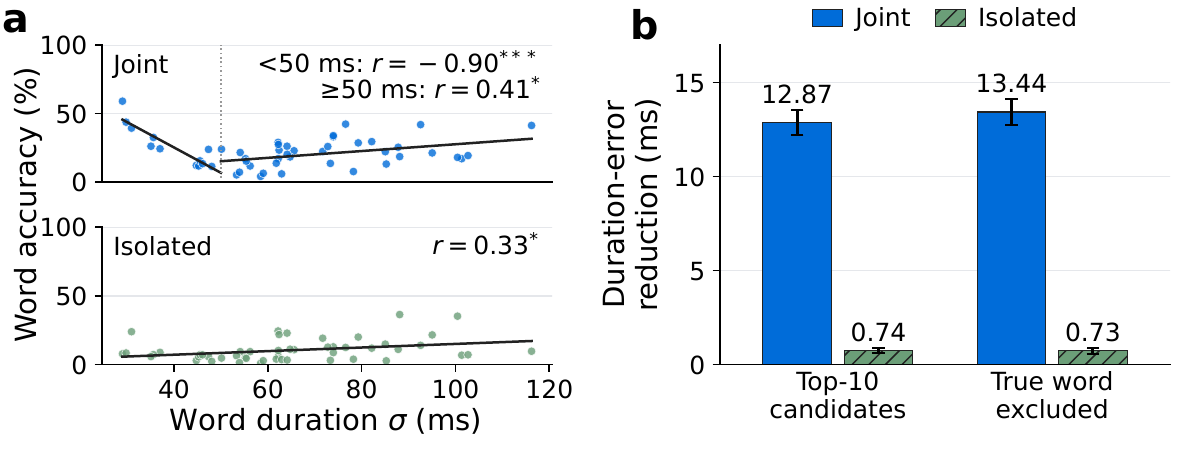}
    \caption{\looseness=-1 \textbf{Jointly decoding words is sensitive to word duration.}
\textbf{(a)} Each point is a word with accuracy plotted against the
standard deviation of its duration across training occurrences.
\textbf{(b)} We test whether the decoder predicts words whose typical durations match the test occurrence. We measure how far the typical durations of the model’s top-10 predictions are from an occurrence’s duration. We then replace the duration with that of another occurrence of the same true word and recompute the error. The bar shows how much the error increases after this replacement, so positive values mean the predictions have durations more similar to the specific occurrence than expected. Error bars show standard
deviation. $^*p<.05$, $^{***}p<.001$.}
    \label{fig:durationdiagnostics}
    \vspace{-1em}
\end{figure}

\looseness=-1 We compare five conditions. \emph{Joint (MEG)} uses the full decoder described
by \citet{dAscoli2025TowardsDI} on real MEG data. \emph{Isolated (MEG)} is similar, except it processes each word window independently in isolation, providing a reference for the gain from jointly decoding words. To test whether this gain requires neural information, in the \emph{Joint (synthetic)} control, we replace MEG
with a continuous synthetic signal that is unrelated to the stimulus, but
extract windows at the same true word onsets. Adjacent synthetic windows therefore share the same underlying samples at offsets determined by the original word onsets. In the \emph{Isolated (synthetic)} control, we generate a separate
synthetic signal for each word window rather than extracting all windows from
one continuous signal. Neighbouring inputs therefore share no samples, so their
relative positions no longer reveal the true intervals between words.
Comparing the synthetic joint and synthetic isolated conditions shows the information
provided by overlapping windows. Lastly, in the \emph{Timing} condition, we encode the log interval between words and supply only this to the same model.

\looseness=-1 \paragraph{Results.} Figure~\ref{fig:controls} shows that the gain from jointly decoding words can be reproduced almost entirely without neural information. On MEG, jointly decoding words increases word accuracy from 9.5\% to 22.3\%. A synthetic
signal with the same overlap structure reaches 22.0\%, despite containing no
brain activity. Both reach similar accuracy to the model trained directly on word intervals as inputs (22.9\%). When this overlap is removed, accuracy falls to 5.8\%. We reproduce this effect across a further 32 subjects (Appendix~\ref{app:multisub}) and across more datasets (Appendix~\ref{app:crossdataset}). We also find that predictions from jointly decoding words align more strongly with duration-based
word probabilities than isolated predictions
(Appendix~\ref{sec:timing-agreement}). As an additional control, we retrain and evaluate on non-overlapping natural sentence constructions, preserving the original
word sequences while removing shared samples. This does not recover an
advantage over isolated word decoding (Appendix~\ref{app:nonoverlap-context}).

Figure~\ref{fig:durationdiagnostics} provides further evidence that jointly decoding words relies on word duration. In panel~a, this model is substantially more accurate for words whose durations are consistent across training occurrences, an effect that is much weaker for isolated decoding. Panel~b shows that this model also favours candidate words whose typical durations match the duration of the specific test occurrence, again much more strongly than the isolated model.

\looseness=-1 These results do not imply that jointly decoding words is inherently unhelpful.
It could still be useful when neighbouring responses can be modelled
jointly without exposing word timing, for example when words are sufficiently
separated in time to avoid overlap. In naturalistic speech, however, avoiding this timing information in word-aligned B2T is difficult because truncating or masking windows at word boundaries can itself reveal the same timing information. In settings where word onsets are unknown, this kind of leakage is not an issue. However, such a setting represents a substantially different and more difficult task, which we do not evaluate here. In the word-aligned B2T setting, decoding words individually presents a way to force a neural network to predict words from brain activity without using this shortcut.

\section{Removing Timing Shortcuts Improves Sentence Reconstruction}
\label{sec:simpleb2t}

Having shown that much of the gain from jointly decoding words can arise from
non-neural word duration information, we next ask what happens when we decode words independently to see how the neural network performs when it does not use the shortcut. We introduce a simple isolated word decoder
and test whether its predictions can be more effectively used to reconstruct
sentences, and whether these predictions can now benefit from established strategies (aggregating observations and using a language model prior).  We test these questions on a clinically motivated communication benchmark and provide additional technical details of all experiments in Appendix~\ref{app:expdetails}.

\subsection{Decoding Words Independently With A Simple Neural Network}

\looseness=-1 We retain the word-level encoder, semantic targets, and contrastive training
framework of \citet{dAscoli2025TowardsDI}, while decoding words independently.

Let $x_i$ denote the three-second MEG window aligned to word $w_i$.
A neural network $f_\theta$ maps this window to a unit-normalised embedding
\begin{equation}
    \hat z_i = f_\theta(x_i).
\end{equation}
Here, $f_\theta$ consists of a small CNN, whose output embeddings are temporally pooled, followed by a residual MLP that outputs $\hat z_i$.
Effectively, $f_\theta$ is a simplification of the \citeauthor{dAscoli2025TowardsDI} encoder, which processes all $x_i$ in a sentence through the same small CNN with temporal pooling, and then jointly processes the pooled embeddings across a sentence with a large 16-layer, 16-head bidirectional transformer. The \citeauthor{dAscoli2025TowardsDI} model has approximately 200 million parameters, while our simplification has about 20 million. Unlike \citeauthor{dAscoli2025TowardsDI}'s encoder,
$f_\theta$ has no access to the
other windows in the sentence. A three-second window can nevertheless contain neural responses to subsequent
words. We examine whether the predictability of this
subsequent context is associated with decoding accuracy in
Appendix~\ref{sec:future-predictability}.

\begin{figure}[t]
    \centering
    \includegraphics[width=1.0\linewidth]{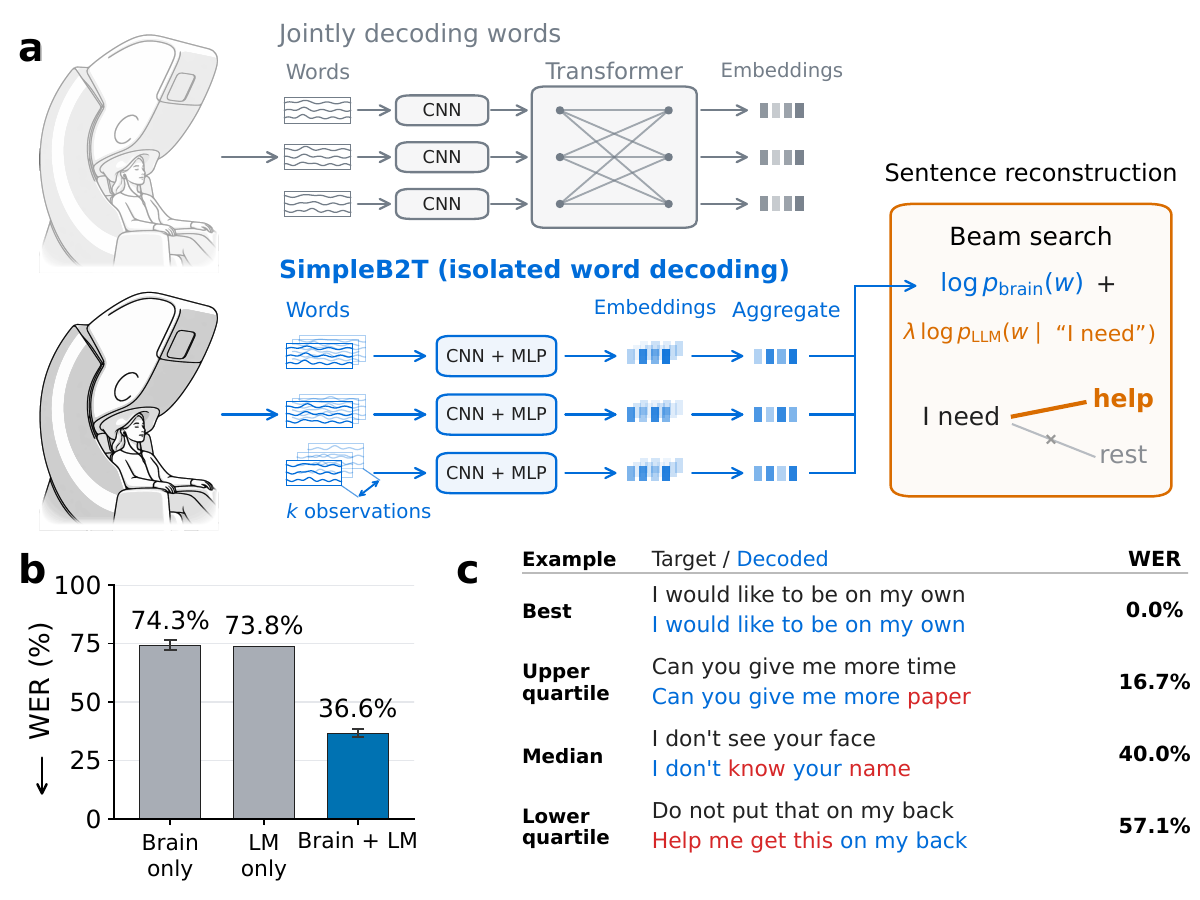}
    \caption{\looseness=-1 \textbf{SimpleB2T.}
\textbf{(a)} Instead of jointly decoding words, we decode each neural response independently. We also test the effect of aggregating responses from distinct observations and using an LLM with beam search for sentence decoding.
\textbf{(b)} Combining predictions with an LLM substantially reduces WER.
\textbf{(c)} Decoded examples. Results are on our core set of 100 clinical sentences with $k=5$ observations per word. Error bars indicate $\pm1$ standard deviation across five training seeds.}
    \label{fig:overview}
    \vspace{-1em}
\end{figure}

As before, $f_\theta$ is trained contrastively, following
\citet{dAscoli2025TowardsDI}, so that $\hat z_i$ is close to $e_{w_i}$ and far
from embeddings of other words. At test time, the cosine similarity
$\hat z_i^\top e_w$ therefore measures the support for candidate word
$w$.

For a candidate vocabulary of words $\mathcal V$, we convert these similarities into a
distribution over words:
\begin{equation}
    p_{\mathrm{brain}}(w \mid x_i)
    =
    \frac{
        \exp(\hat z_i^\top e_w / T)
    }{
        \sum_{v \in \mathcal V}
        \exp(\hat z_i^\top e_v / T)
    }.
    \label{eq:lexical_probability}
\end{equation}
The temperature $T$ is selected once on validation data and then fixed.
This step turns the retrieval scores of the neural decoder into
probabilities that can later be combined with a language model.

\subsection{Combining Observations and Linguistic Priors}

Since non-invasive neural responses have limited signal-to-noise ratio,
we also test two simple ways to strengthen word predictions for sentence
reconstruction. Firstly, when multiple responses to the same word are available,
we aggregate their predictions. Secondly, we combine the resulting neural
probabilities with an off-the-shelf language model, providing a linguistic prior to favour plausible word sequences. We refer to the recipe of independently decoding words, using aggregation, and leveraging an LLM prior as \emph{SimpleB2T}. We provide further technical details in Appendix~\ref{app:details}.

\paragraph{Aggregating observations.}
When $k$ distinct observations of the same word are available, we encode each
independently and average their predicted embeddings,
\begin{equation}
    m_i = \frac{1}{k}\sum_{j=1}^k \hat z_{i,j},
    \qquad
    r_i = \|m_i\|_2,
    \qquad
    u_i = \frac{m_i}{r_i}.
\end{equation}
As each predicted embedding is unit normalised, $r_i$ measures agreement
between observations and approaches one when their predictions point in similar
directions and decreases when they disagree. For candidate word $w$,
we define the aggregated neural score
\begin{equation}
    A_i(w)
    =
    \frac{1+\alpha r_i}{T}\,u_i^\top e_w,
    \label{eq:aggregation_score}
\end{equation}
where the agreement weight $\alpha$ controls how strongly agreement between observations influences
the score. Normalising $m_i$ to obtain the consensus direction $u_i$ would discard this information, so we use $r_i$ to modulate the sharpness of the neural scores. 

\paragraph{LLM rescoring.}
For a candidate sentence
$y=(w_1,\ldots,w_n)$, we combine the neural scores with the autoregressive
likelihood assigned by a language model:
\begin{equation}
    S(y)
    =
    \sum_{i=1}^{n} A_i(w_i)
    +
    \lambda \log p_{\mathrm{LLM}}(y \mid q),
    \label{eq:sentence_score}
\end{equation}
where $q$ is a fixed prompt and $\lambda$ controls the contribution of the
language model. In this paper, $p_{\mathrm{LLM}}$ is the base model of Qwen3-8B \citep{yang2025qwen3}. We approximately maximise \Cref{eq:sentence_score} using
left-to-right beam search over the candidate vocabulary. As in
\citet{dAscoli2025TowardsDI}, the number of word positions is known for each
test sentence, so decoding terminates after that many words. This is part of the word-aligned B2T setting.

\subsection{Experimental Setup}
\looseness=-1 We largely follow the experimental setup described in Section~\ref{sec:timing_audit}. For this evaluation, however, validation and test decoding are
performed over a vocabulary of 92 words on the sentences in our clinical communication benchmark, described next. We report WER and the proportion of sentences where all words are decoded correctly, which we call the \emph{sentence match rate (SMR)}.

\looseness=-1 \paragraph{Clinical communication benchmark.} Our primary evaluation is a benchmark of 200 clinically motivated
communication sentences constructed from our held-out test recordings.
The sentences are designed to represent simple patient communication needs such as
requesting help, water, repositioning, environmental changes, or interaction
with another person. They are divided into two 100-sentence subsets:
\emph{Core}, containing short care-related
utterances, and \emph{Expanded}, containing a broader range of
questions, statements, temporal expressions, and social interactions. These clinically motivated sentences are inspired by the task design of the surgical decoding work from \citet{Moses2021NeuroprosthesisFD}, who used a smaller 50-word vocabulary and 50 total sentences. This constrained communication domain also provides a stronger linguistic
prior, helping an LLM resolve uncertain neural predictions into plausible
patient messages. Each sentence position is constructed using a recorded occurrence
of the corresponding word from the held-out test sessions. Consequently,
neighbouring sentence positions come from distinct recording events and do not
share overlapping signal samples, so the shortcut exploited by
jointly decoding words is unavailable. To allow for aggregating neural responses, each sentence position is assigned
up to five distinct occurrences of the same word. Occurrences are sampled
without replacement and are not reused across benchmark positions. Thus, five
occurrences correspond to five separately recorded MEG responses to the same
word. A list of all sentences is provided in Appendix~\ref{app:benchmark}.

\subsection{Results}

\begin{table*}[t]
\centering
\footnotesize
\setlength{\tabcolsep}{3pt}
\renewcommand{\arraystretch}{1.05}
\begin{tabular*}{\textwidth}{@{\extracolsep{\fill}}lc*{6}{c}@{}}
 & & \multicolumn{2}{c}{Core} & \multicolumn{2}{c}{Expanded} & \multicolumn{2}{c}{Full} \\
\cmidrule(lr){3-4}\cmidrule(lr){5-6}\cmidrule(lr){7-8}
Method & $k$ & WER (\%) $\downarrow$ & \shortstack{SMR (\%) $\uparrow$} & WER (\%) $\downarrow$ & \shortstack{SMR (\%) $\uparrow$} & WER (\%) $\downarrow$ & \shortstack{SMR (\%) $\uparrow$} \\
\midrule
LM only & - & $73.8$ & $2.0$ & $91.8$ & $0.0$ & $83.4$ & $1.0$ \\
\midrule
d'Ascoli & 1 & $98.9_{\pm0.3}$ & $0.0_{\pm0.0}$ & $98.8_{\pm0.1}$ & $0.0_{\pm0.0}$ & $98.8_{\pm0.2}$ & $0.0_{\pm0.0}$ \\
d'Ascoli + LM & 1 & $77.3_{\pm1.1}$ & $2.4_{\pm0.3}$ & $87.5_{\pm0.8}$ & $0.2_{\pm0.1}$ & $82.7_{\pm0.2}$ & $1.3_{\pm0.2}$ \\
SimpleB2T & 1 & \cellcolor{gray!15}$\mathbf{65.6_{\pm0.8}}$ & \cellcolor{gray!15}$\mathbf{6.4_{\pm0.5}}$ & \cellcolor{gray!15}$\mathbf{76.3_{\pm0.5}}$ & \cellcolor{gray!15}$\mathbf{4.2_{\pm0.4}}$ & \cellcolor{gray!15}$\mathbf{71.3_{\pm0.5}}$ & \cellcolor{gray!15}$\mathbf{5.3_{\pm0.3}}$ \\
\midrule
d'Ascoli & 5 & $98.9_{\pm0.3}$ & $0.0_{\pm0.0}$ & $98.8_{\pm0.4}$ & $0.0_{\pm0.0}$ & $98.9_{\pm0.1}$ & $0.0_{\pm0.0}$ \\
d'Ascoli + LM & 5 & $79.4_{\pm2.5}$ & $2.4_{\pm0.5}$ & $85.5_{\pm2.0}$ & $0.2_{\pm0.4}$ & $82.7_{\pm0.6}$ & $1.3_{\pm0.3}$ \\
SimpleB2T & 5 & \cellcolor{gray!15}$\mathbf{36.6_{\pm1.7}}$ & \cellcolor{gray!15}$\mathbf{26.0_{\pm2.5}}$ & \cellcolor{gray!15}$\mathbf{44.8_{\pm3.7}}$ & \cellcolor{gray!15}$\mathbf{20.2_{\pm4.3}}$ & \cellcolor{gray!15}$\mathbf{41.0_{\pm2.2}}$ & \cellcolor{gray!15}$\mathbf{23.1_{\pm3.0}}$ \\
\end{tabular*}
\caption{\textbf{Effect of decoding words independently.}
We compare SimpleB2T and
\citeauthor{dAscoli2025TowardsDI}, with and without the same LLM rescoring. The table tests how using $k$ observations and LLM rescoring behave
when words are decoded jointly and when words are decoded independently. Results are means across five seeds and subscripts show sample standard deviations.}
\label{tab:clinical-comparison}
\end{table*}

\begin{table*}[t]
\centering
\setlength{\tabcolsep}{2pt}
\renewcommand{\arraystretch}{1.05}
\captionsetup[subtable]{font=footnotesize,labelfont=normalfont,justification=raggedright,singlelinecheck=false,skip=2.5pt}
\begin{subtable}[t]{0.30\textwidth}
\vspace{0pt}\centering
\footnotesize
\begin{tabular}{@{}
>{\raggedright\arraybackslash}p{\dimexpr0.65\linewidth-2\tabcolsep\relax}
>{\centering\arraybackslash}p{0.35\linewidth}
@{}}
LM & WER (\%) \\
\midrule
Without & $74.3_{\pm2.0}$ \\
With & \cellcolor{gray!15}$\mathbf{36.6_{\pm1.7}}$ \\
\multicolumn{2}{c}{\strut} \\
\multicolumn{2}{c}{\strut} \\
\end{tabular}
\caption{\textbf{LM contribution.} LM rescoring halves WER.}
\label{tab:clinical-ablation-a}
\end{subtable}
\hfill
\begin{subtable}[t]{0.30\textwidth}
\vspace{0pt}\centering
\footnotesize
\begin{tabular}{@{}
>{\raggedright\arraybackslash}p{\dimexpr0.65\linewidth-2\tabcolsep\relax}
>{\centering\arraybackslash}p{0.35\linewidth}
@{}}
Input & WER (\%) \\
\midrule
LM only & $73.8$ \\
Noise + LM & $97.1_{\pm1.7}$ \\
Shuff. brain + LM & $88.3_{\pm1.9}$ \\
Brain + LM & \cellcolor{gray!15}$\mathbf{36.6_{\pm1.7}}$ \\
\end{tabular}
\caption{\textbf{Controls.} Results depend on example-specific information.}
\label{tab:clinical-ablation-b}
\end{subtable}
\hfill
\begin{subtable}[t]{0.30\textwidth}
\vspace{0pt}\centering
\footnotesize
\begin{tabular}{@{}
>{\raggedright\arraybackslash}p{\dimexpr0.65\linewidth-2\tabcolsep\relax}
>{\centering\arraybackslash}p{0.35\linewidth}
@{}}
Prompt $q$ & WER (\%) \\
\midrule
Generic & $55.2_{\pm1.2}$ \\
Task-oriented & \cellcolor{gray!15}$\mathbf{36.6_{\pm1.7}}$ \\
\multicolumn{2}{c}{\strut} \\
\end{tabular}
\caption{\textbf{LM prompt.} Describing the task and domain reduces error.}
\label{tab:clinical-ablation-c}
\end{subtable}
\par\vspace{10pt}
\begin{subtable}[t]{0.30\textwidth}
\vspace{0pt}\centering
\footnotesize
\begin{tabular}{@{}
>{\raggedright\arraybackslash}p{\dimexpr0.65\linewidth-2\tabcolsep\relax}
>{\centering\arraybackslash}p{0.35\linewidth}
@{}}
Beam width & WER (\%) \\
\midrule
1 & $67.7_{\pm1.6}$ \\
25 & $37.9_{\pm2.4}$ \\
50 & \cellcolor{gray!15}$\mathbf{36.6_{\pm1.7}}$ \\
200 & ${35.9_{\pm2.5}}$ \\
500 & $36.0_{\pm2.7}$ \\
\end{tabular}
\caption{\textbf{Beam width.} WER plateaus beyond width 200.}
\label{tab:clinical-ablation-d}
\end{subtable}
\hfill
\begin{subtable}[t]{0.30\textwidth}
\vspace{0pt}\centering
\footnotesize
\begin{tabular}{@{}
>{\raggedright\arraybackslash}p{\dimexpr0.65\linewidth-2\tabcolsep\relax}
>{\centering\arraybackslash}p{0.35\linewidth}
@{}}
LM weight $\lambda$ & WER (\%) \\
\midrule
0 & $74.3_{\pm2.0}$ \\
0.25 & $40.8_{\pm2.6}$ \\
0.5 & \cellcolor{gray!15}$\mathbf{36.6_{\pm1.7}}$ \\
1 & $47.7_{\pm1.3}$ \\
2 & $59.7_{\pm0.8}$ \\
\end{tabular}
\caption{\textbf{LM weight.} LM evidence is most useful at half-weight.}
\label{tab:clinical-ablation-e}
\end{subtable}
\hfill
\begin{subtable}[t]{0.30\textwidth}
\vspace{0pt}\centering
\footnotesize
\begin{tabular}{@{}
>{\raggedright\arraybackslash}p{\dimexpr0.65\linewidth-2\tabcolsep\relax}
>{\centering\arraybackslash}p{0.35\linewidth}
@{}}
Agreement wt. $\alpha$ & WER (\%) \\
\midrule
0 & $49.3_{\pm1.5}$ \\
0.5 & $42.7_{\pm1.0}$ \\
1 & $40.7_{\pm1.5}$ \\
2 & \cellcolor{gray!15}$\mathbf{36.6_{\pm1.7}}$ \\
4 & $38.1_{\pm2.6}$ \\
\end{tabular}
\caption{\textbf{Agreement weight.} Weight 2 gives the lowest WER.}
\label{tab:clinical-ablation-f}
\end{subtable}
\caption{\textbf{Ablations.} Values selected on a development set of 50 sentences, except for (c) which is retrospective and did not inform $q$. Results on the core set of test sentences with $k=5$. \textbf{Bold} indicates the selected value. Results are means across training seeds, with sample standard deviations.}
\label{tab:clinical-ablations}
\vspace{-1em}
\end{table*}

\looseness=-1 When decoding words jointly, aggregation provides little benefit, and LM rescoring yields performance close to or worse than the LM-only baseline (Table~\ref{tab:clinical-comparison}).
After decoding words independently, both become substantially more effective. A likely explanation on our benchmark is that the shortcut can no longer be exploited, so predictions from \citeauthor{dAscoli2025TowardsDI}'s model provide little information beyond the LLM. More generally, predictions from jointly decoding words may not have benefitted from linguistic priors because the information output by a decoder that learns the shortcut partly reflects word duration statistics, which may be less complementary to an external LLM prior. We analyse a version of the model retrained without overlapping windows in Appendix~\ref{app:traintestbench}, where we continue to find no benefit from jointly decoding words.
By contrast,
isolated word decoding must make each prediction from a single neural response,
producing information from brain activity that can be combined more effectively across observations
and with an LLM. At $k=5$, LLM rescoring reduces WER from 74.3\% to 36.6\%,
while increasing the number of observations from one to five reduces WER from
65.6\% to 36.6\%.

\begin{figure}[t]
    \centering
    \includegraphics[width=1.0\linewidth]{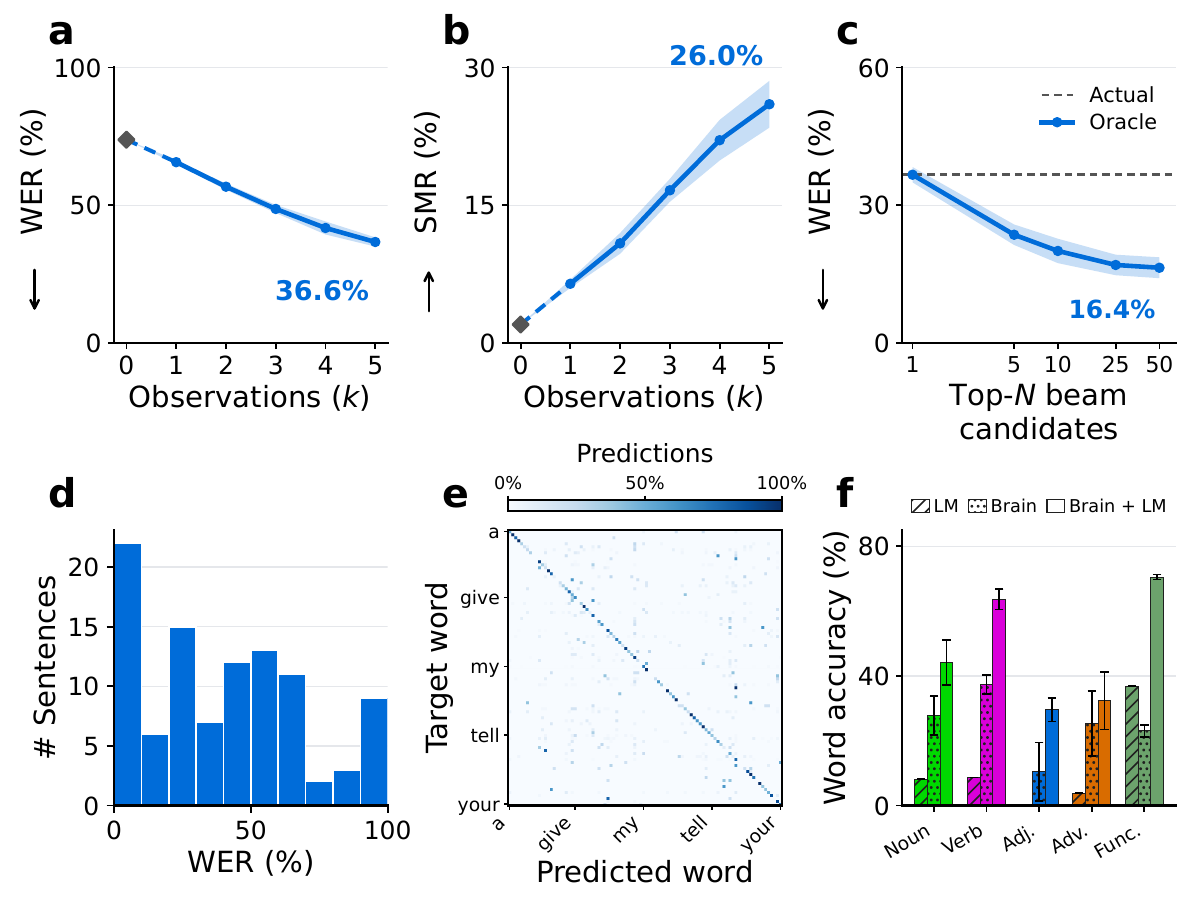}
    \caption{\looseness=-1 \textbf{SimpleB2T results.} 
    \textbf{\textbf{(a, b)}} WER and exact sentence match as the number of observations increases. 
    \textbf{(c)} Oracle selection among the top-$N$ beam candidates.
    \textbf{(d)} Sentence WER distribution.
    \textbf{(e)} Confusion matrix across the 92-word vocabulary.
    \textbf{(f)} Word accuracy by part of speech for the LM, neural decoder, and their combination.
    Panels c--f use $k=5$; all results are on the Core test set.
Shading and error bars indicate $\pm1$ standard deviation across five training seeds.}
    \label{fig:results}
    \vspace{-1em}
\end{figure}

\paragraph{Aggregating observations provides a trade-off between measurement and accuracy.}
Figures~\ref{fig:results}a--b show that sentence decoding improves steadily as
additional responses to the same word are combined, without retraining the
neural decoder. Thus, collecting more observations provides a simple way to trade
additional recording burden for more reliable predictions. We also note that the language-model weight selected on the development set decreases with $k$, consistent with the neural predictions becoming more informative as observations are combined (Appendix~\ref{app:weight}).

\looseness=-1 \paragraph{Neural and linguistic information are complementary.} At $k=5$, the neural decoder obtains 74.3\% WER and the LLM alone 73.8\%. Combining them reduces WER to 36.6\% (\Cref{tab:clinical-ablations}). This improvement depends on information specific to the corresponding brain response. Noise inputs matching the statistics of MEG give 97.1\% WER
and shuffling the predicted embeddings across test occurrences gives 88.3\%. The
complementarity also shows across parts of speech
(Figure~\ref{fig:results}f), where neural evidence helps verbs
while the LLM contributes strongly to function words. Additional analyses are provided in Appendix~\ref{app:word-analysis}.

\paragraph{Errors may be reduced without changing the neural decoder.}
At $k=5$, better sentences are often already present within the
beam than the sentence ultimately selected (Figure~\ref{fig:results}c).
This suggests that the neural decoder and language model often recover enough
information to generate good reconstructions, but the current scoring rule does
not always rank them correctly. The linguistic prior itself is also important.
Explicitly describing the patient communication setting substantially improves
decoding (Table~\ref{tab:clinical-ablation-c}; Appendix~\ref{app:prior}),
presumably because it concentrates probability on the restricted set of messages
that are plausible in this setting and therefore helps resolve ambiguous neural
predictions. These results suggest that an important remaining
bottleneck is how neural evidence and linguistic context are used to select
among plausible sentences.

\section{Discussion}

\looseness=-1 Much of the gain from jointly decoding words reported by \citet{dAscoli2025TowardsDI} can be reproduced without brain information. Improvements arise because overlapping word-aligned windows expose word durations, which are informative about word identity. Therefore, when providing external information, one should be careful not to enable artificial shortcuts. To avoid the shortcut here, we decode words independently instead of jointly and find that two established neural decoding techniques---combining
multiple observations and using an explicit linguistic prior---become
substantially more effective. When predictions are driven by timing, aggregating observations gives a better estimate of a word’s typical duration, improving predictions through the same shortcut. Without the shortcut, aggregation instead improves the estimate of the word-specific neural signal. Similarly, this signal is more complementary to LLM priors than information from the shortcut.

\looseness=-1 \paragraph{When does the shortcut arise?}
The shortcut found here occurs when jointly decoding words where the interval between word onsets vary. This is common in natural speech datasets and in reading paradigms where presentation timing varies by word. Of the nine datasets analysed in \citet{dAscoli2025TowardsDI}, only LittlePrinceRead and Nieuwland are unaffected as they use reading protocols that provide the same amount of time to each word. Intriguingly, these are also the only two datasets for which the authors find that jointly decoding words does not improve performance compared to decoding words independently. This is consistent with our own finding that training a model to jointly decode words without the shortcut does not improve results (Appendix~\ref{app:traintestbench}). 

\looseness=-1 \paragraph{Scope of the timing-leakage result.}
Since its publication in late 2025, the approach introduced by
\citeauthor{dAscoli2025TowardsDI} has already been extended by several later
brain-to-text methods, e.g. \citet{jayalath2025unlocking}, \citet{Lvy2026NoninvasiveDO}, \citet{wang2026margin}, and \citet{li2026hdnd}, broadening the issue beyond the original work. This method has also been evaluated in several subsequent studies \citep{zhang2025thought, jayalath2026meg, ozdogan2025libribrain, mantegna2026pnplcompetition, landau2026semantic}, included in benchmarks \citep{banville2026neuralbench, jayalath2026common}, and discussed in recent reviews \citep{yang2026review}. Brain2Qwerty \citep{Lvy2026NoninvasiveDO} is distinct in that it operates on character-aligned windows from typing, potentially providing an analogous timing shortcut for keystroke intervals. However, our synthetic control suggests that timing explains little of its performance (Appendix~\ref{app:b2q}). Brain2Qwerty v2 \citep{zhang2026accurate}, by contrast, does not assume aligned windows. 
The timing shortcut is therefore relevant to a growing line of non-invasive brain-to-text work, although its importance depends on how informative timing is in each task. More generally, analogous shortcuts could arise whenever aligned, overlapping inputs are modelled jointly.

\looseness=-1 \paragraph{Repeated observations and word alignment.}
Our evaluation assumes that the temporal location of each word is known (word-aligned B2T), following \citet{dAscoli2025TowardsDI}. This differs from solving speech segmentation jointly with word decoding, but is not uncommon among existing BCIs: for example, \citet{Moses2021NeuroprosthesisFD} used a separate
neural speech-detection model to identify the onset and offset of attempted
words before classification. A non-invasive system could similarly
use an interface in which individual word attempts are made separately,
making segmentation substantially easier with non-invasive speech detection models \citep{Dash2020NeuroVADRV, bbljayalath2025}. Such a setting would also make gathering
multiple observations natural by repeating attempts. Repetition is already widely used in simple
non-invasive BCIs such as P300 spellers, where multiple responses are combined to improve signal-to-noise ratio
\citep{Farwell1988TalkingOT}. The scale of this effect here is notable relative to comparable speech decoding work. In the 2025 PNPL competition \citep{landau2025competition}, averaging 100 observations per
phoneme was used to improve decoding reliability, whereas SimpleB2T obtains
large sentence-level gains from only five observations per word.

\looseness=-1 \paragraph{Towards useful non-invasive speech decoding.}
In 2021, \citeauthor{Moses2021NeuroprosthesisFD} provided a landmark
demonstration that attempted speech could be decoded into sentences in a
person with paralysis, achieving a median WER of 25.6\% using a 50-word
vocabulary and a set of 50 test sentences. While we do not decode speech from paralysed patients, and have not yet solved the problem of segmenting words without knowing their onsets, SimpleB2T shows that in decoding perceived speech from non-invasive recordings, it is possible to achieve WERs much closer
to that landmark invasive result. This is possible while using multiple observations and operating over a larger
92-word vocabulary within a broader communication benchmark. Like their work, we focus on a constrained communication setting which allows for a stronger linguistic prior than unrestricted language decoding.

\looseness=-1 Once independent word decoding removes the timing shortcut introduced by the word-aligned input construction, the remaining challenges become clearer. 
Better selection among the top beam search candidates and reducing the number of observations required are opportunities for future work.
More importantly, extending these results to decoding internally generated speech, such as imagined speech, will be a key step towards practical non-invasive communication interfaces.

\subsubsection*{Acknowledgments}

We thank Gilad Landau, Francesco Mantegna, and Yonatan Gideoni for helpful comments. We also thank Stéphane d'Ascoli for his correspondence and for checking a draft of this work.

We are grateful to \href{https://modal.com}{Modal Labs, Inc.} for a generous compute grant which helped support this project. In particular, we thank Adam Azzam for extending the grant timeline to meet conference deadlines.

\clearpage
\subsection*{AI use statement}

We used generative AI tools to assist with construction of the clinically
motivated communication benchmark and to assist with
drafting and editing the manuscript. All generated benchmark sentences were
reviewed by the authors before use. Generative AI was also used to assist with code generation and debugging. All resulting code was inspected and its outputs were verified by the authors. All experimental designs, analyses, results, and scientific
claims were reviewed and verified by the authors. We take responsibility for
the final content of the paper, including text and artifacts produced with the
aid of generative AI.

\subsection*{Ethics statement}

This work involves no new human-subject data collection. All neural recordings
come from the publicly released LibriBrain100 dataset where ethical approval and
participant consent for the original data collection are described by
\citet{mantegna2026}. The communication benchmark we constructed from this data is not clinically validated. Thus, we encourage user-centred evaluation before any real-world use. More broadly, progress
in neural speech decoding raises important questions about privacy and consent. Practical systems should only decode deliberately provided signals
with the informed consent of their users. However, we do not interpret the reported method as yet demonstrating a practical BCI since it operates on perceived and not imagined speech.

\subsection*{Reproducibility statement}

We provide the information required to reproduce the experiments throughout
the paper and appendices. Section~\ref{sec:leakage} describes the protocol for our controls testing the overlap-window timing leakage; Section~\ref{sec:simpleb2t} defines SimpleB2T and its scoring procedure;  Appendix~\ref{app:details} gives
the preprocessing, architecture, training, and language-model decoding
settings; Appendix~\ref{app:expdetails} specifies the experimental controls; Appendix~\ref{app:data} lists the exact
train, validation, and test splits; and Appendix~\ref{app:benchmark} gives the 200-sentence
communication benchmark and its construction procedure. We report the random seeds used for
training and evaluate models across up to five seeds.

\bibliography{iclr2027_conference}
\bibliographystyle{iclr2027_conference}

\clearpage
\appendix
\addcontentsline{toc}{part}{Appendix}
\etocsetnexttocdepth{subsection}
\etocsettocstyle{\section*{Appendix Contents}}{}
\localtableofcontents
\newpage

\section{Timing-Shortcut Analyses}

\subsection{Relationship Between Word Timing and Duration}
\label{app:timing_sanity}

\begin{figure}
    \centering
    \includegraphics[width=0.65\linewidth]{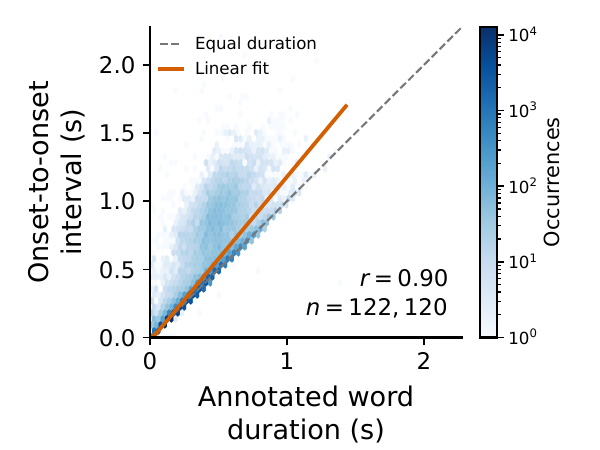}
    \caption{\textbf{Intervals between word onsets closely track word duration.}
Relationship between the annotated duration of a word and the interval from
its onset to the onset of the following word across adjacent word
pairs. The two quantities are strongly correlated ($r=0.90$). The dashed line
indicates equality and the solid line shows a linear fit.}
    \label{fig:duration_interval}
\end{figure}

\looseness=-1 The timing shortcut we identify relies on two properties of the word-aligned construction, namely that
neighbouring windows must overlap, and the resulting onset-to-onset intervals
must contain information about word duration. We find that both hold strongly in
the perceived speech data we study. Three-second windows overlap for 99.9996\% of adjacent word
pairs.
Across 122,120 adjacent word pairs, the onset-to-onset interval is strongly
correlated with the annotated duration of the preceding word
($r=0.90$; Figure~\ref{fig:duration_interval}).

\subsection{Synthetic Control Details}

\paragraph{Joint synthetic signal.}
To test whether a contextual decoder can use window overlap without any
neural information, we replace the MEG with a synthetic continuous signal.
A simple Gaussian-noise signal would preserve identical samples across
overlapping windows, but provides little structured variation for the CNN to
encode. We therefore generate a sparse, smoothly varying signal with
recognisable local features whose positions can be recovered after a temporal
shift.

Specifically, we generate a 306-channel signal sampled at 50~Hz from three
independent pulse processes, smoothed with Gaussian kernels of standard
deviation 60, 120, and 240~ms. Pulses occur at a combined rate of 3~Hz.
Pulse locations are shared across channels, while their amplitudes are sampled
independently for each channel. The three smoothed components are summed and
normalised to unit population variance. Importantly, pulse generation is
entirely independent of the words and their onset times.

For each sentence, we generate one continuous synthetic signal and extract
3-second windows at the \emph{true} word onsets, using the same windowing
procedure as for the real MEG. Neighbouring windows therefore contain the same
synthetic features at relative offsets determined by the true inter-word
timing. Each window is baseline-corrected using its first 0.5~s and clipped to
$[-5,5]$.

\paragraph{Isolated synthetic signal.}
The independent control uses the same synthetic signal generator, but samples
a separate signal for every word window. Consequently, neighbouring windows
share no underlying samples and the true relative word timings are no longer
encoded in the inputs.

During training, synthetic signals are regenerated on every forward pass.
Validation and test signals are generated deterministically using
sentence-specific random seeds. Comparing the shared and independent
conditions therefore isolates the information provided by overlap between
word-aligned windows.

\paragraph{Timing.} To test whether word intervals alone support decoding, we train a
timing-only control that replaces each MEG window with the interval
to the next word onset within the same sentence. Intervals are log-transformed and standardised using training-set statistics. A two-layer MLP maps each scalar to a 1024-dimensional,
$\ell_2$-normalised representation, which is passed to the same
sentence-level transformer and trained with the same contrastive
objective.

\subsection{Timing Shortcut Reproduction Across Multiple Subjects}
\label{app:multisub}

\begin{figure}
    \centering
    \includegraphics[width=1.0\linewidth]{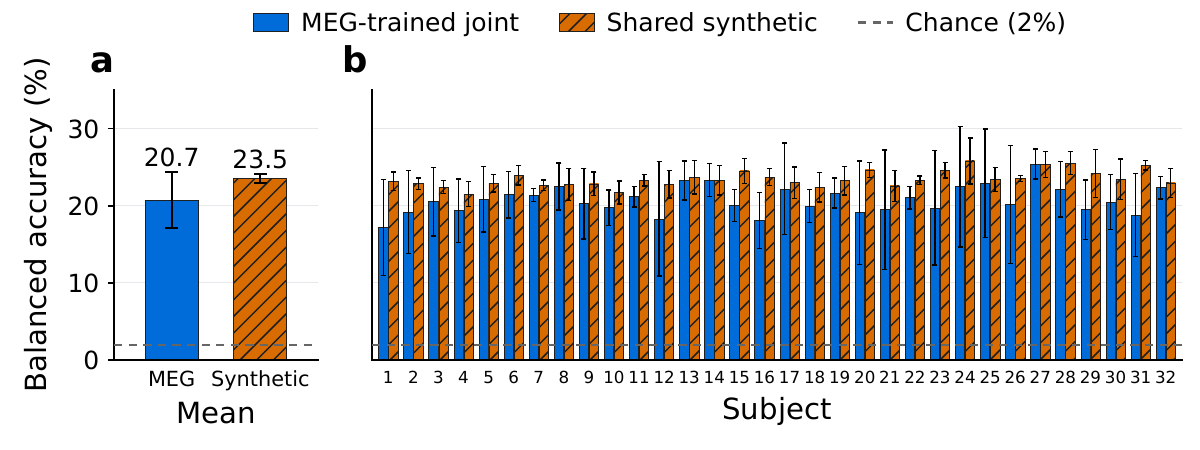}
    \caption{\textbf{The overlapping window shortcut reproduces across all subjects.}
Subject-0-trained MEG and shared-synthetic decoders evaluated on the 32 other subjects in LibriBrain100.
\textbf{(a)} Mean across subjects.
\textbf{(b)} Per-subject results.
Error bars show standard deviation across five training seeds.}
    \label{fig:multisub}
\end{figure}

To test whether the overlapping-window shortcut generalises beyond the subject used for training, we evaluate the same checkpoint on all 32 remaining LibriBrain100 subjects. As shown in Figure~\ref{fig:multisub}, the shared synthetic decoder remains highly accurate across subjects despite receiving no neural information, and achieves performance comparable to the MEG-trained decoder. This shows that the shortcut is not specific to subject 0 and the overlap structure alone provides a transferable signal that can be exploited across unseen subjects and unseen speech material.

\subsection{Timing Shortcut Reproduction Across Datasets}
\label{app:crossdataset}

\begin{figure}
    \centering
    \includegraphics[width=\linewidth]{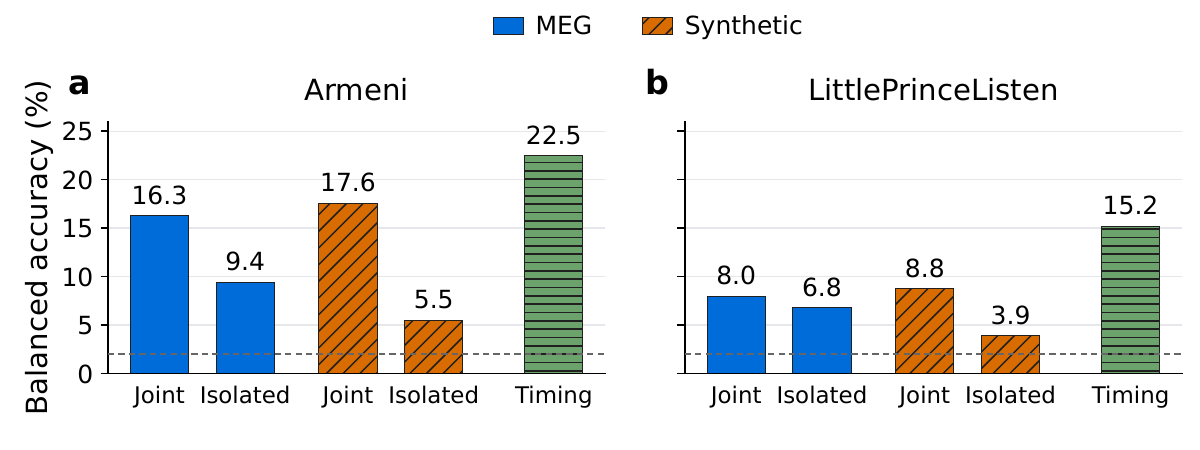}
    \caption{\textbf{Replication of Figure~\ref{fig:controls}b} with \citet{armeni_10-hour_2022} and the listening component of Le Petit Prince MEG \citep{dAscoli2025TowardsDI}.}
    \label{fig:armenicontrols}
\end{figure}

Figure~\ref{fig:armenicontrols} shows that Figure~\ref{fig:controls}b replicates on other datasets. Specifically, we conduct the same experiment on the perceived speech dataset by \citet{armeni_10-hour_2022} and Le Petit Prince MEG \citep{dAscoli2025TowardsDI}. For \citeauthor{armeni_10-hour_2022}, we use all subjects and sessions 1--8 for training, session 9 for validation, and session 10 for test. For Le Petit Prince, we use the listening component with all subjects and split by book section. We use runs 1--6 for training, 7 for validation, and 8--9 for testing. On both datasets, the synthetic control performs on par with using real MEG data.

\subsection{Agreement with Duration-Based Word Predictions}
\label{sec:timing-agreement}

To examine whether jointly decoding words captures timing information, we measure how closely its predicted word probabilities align with those
inferred from word durations alone. We approximate duration by the
interval between consecutive word onsets within a sentence. For each word in the 50-word vocabulary, we fit a
Gaussian distribution over log intervals using training occurrences.
Combining these likelihoods with training word-frequency priors gives
a duration-based posterior $p_{\mathrm{dur}}(w\mid d)$ where $w$ is a word type and $d$ is the duration in samples.

We compare this posterior with predictions from the joint MEG
decoder, the joint synthetic decoder, and our isolated MEG
decoder. Each decoder's temperature is calibrated by minimising
word-level negative log likelihood on validation data.
For each training seed, we compute the change in Jensen--Shannon divergence
\begin{equation}
\Delta_{\mathrm{JS}}
=
\frac{1}{M}\sum_{i=1}^{M}
\left[
\operatorname{JS}\!\left(p_i,q_{\pi(i)}\right)
-
\operatorname{JS}\!\left(p_i,q_i\right)
\right],
\end{equation}
where $p_i$ is the neural word distribution, $q_i$ is the duration-based
posterior, and $\pi$ is a random permutation of test occurrences.
Shuffling preserves the collection of duration-based predictions while breaking their correspondence with individual inputs.

\begin{figure}
    \centering
    \includegraphics[width=0.65\linewidth]{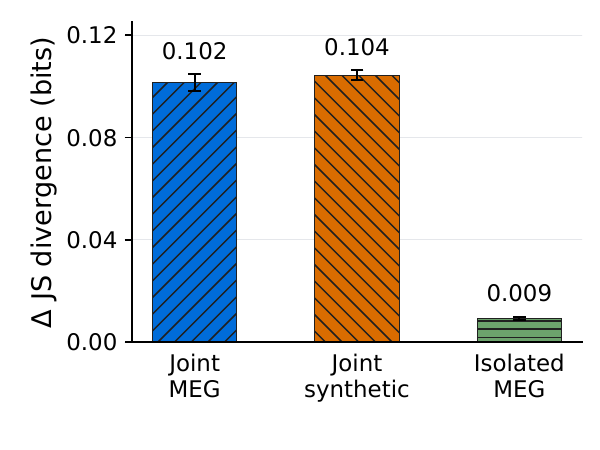}
    \caption{\textbf{Joint decoders aligns with duration-based predictions.}
We measure how much the Jensen--Shannon divergence between the decoder probabilities and a duration-based word predictor's probabilities increase when their pairing across test occurrences is randomly shuffled. Thus, larger values indicate stronger alignment with the duration-based word predictor. Bars show means across five training seeds and error
bars indicate sample standard deviations.}
    \label{fig:timingprob}
\end{figure}

The increase in divergence is substantially larger for the joint
MEG and synthetic decoders
than for the isolated decoder (Figure~\ref{fig:timingprob}).
Thus, both joint decoders exhibit stronger
alignment with evidence derived from the duration of words.
The similar effect for synthetic inputs supports the interpretation
that the overlap structure contributes to performance from jointly decoding words.

\subsection{Jointly Decoding Words Without Overlapping Inputs}
\label{app:nonoverlap-context}

The synthetic signal experiment in Section~\ref{sec:timing_audit} shows that
overlapping word windows are sufficient to reproduce most of the gain from
jointly decoding words without neural information. We perform an additional
control to test whether this result can instead be explained by a train--test
mismatch when overlap is removed.

\looseness=-1 We preserve the original word sequence of each natural sentence, but replace
every position with a separately recorded occurrence of the same word. Any two
windows presented in the same sentence must either come from different
recordings or begin at least three seconds apart, ensuring that they share no
MEG samples. This preserves the linguistic structure of the natural sentences
while removing the overlap between neighbouring neural inputs. We construct new versions of our
training, validation, and test sets in this way independently where any neural response to a word may only be used once. Sentences for which we could not avoid overlaps were discarded. We were able to retain 91\% of all the original sentences in the test recordings and 96\% of the original sentences in the train recordings.

We compare three models: \citet{dAscoli2025TowardsDI}'s decoder trained on overlapping
natural sentences; the same architecture retrained on
non-overlapping sentences; and the isolated neural decoder used by SimpleB2T. All models are evaluated with one observation
per word using balanced top-1 accuracy over the same 50-word vocabulary as in
Section~\ref{sec:timing_audit}.

\begin{figure}[t]
    \centering
    \includegraphics[width=0.65\linewidth]{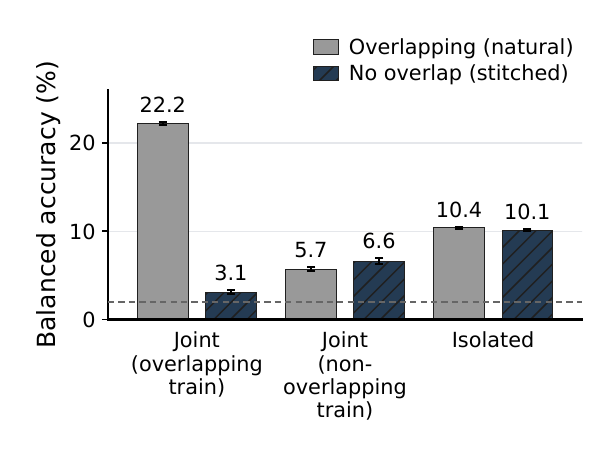}
    \caption{\textbf{Jointly decoding words with and without overlapping inputs.}
    We evaluate \citet{dAscoli2025TowardsDI}, the same architecture trained
    without overlapping neighbouring windows, and isolated decoding on both
    natural overlapping sentences and non-overlapping versions of the same
    word sequences. Error bars indicate $\pm1$ sample standard
    deviation across training seeds.}
    \label{fig:nonoverlap-context}
\end{figure}

Figure~\ref{fig:nonoverlap-context} shows that removing overlap at test time causes the original model to collapse, but this alone could reflect a train--test mismatch. Retraining the same architecture without overlap recovers some performance, yet still does not outperform isolated decoding. Constructing
non-overlapping sentences from separate recordings also removes coherent neural
information shared across neighbouring words, leaving the joint decoding model
primarily able to exploit regularities in the word sequence (effectively acting as an implicit language model).

\subsection{Timing Shortcut Control for Brain2Qwerty}
\label{app:b2q}

Brain2Qwerty~\citep{Lvy2026NoninvasiveDO} also jointly processes aligned neural
windows, raising the possibility of an analogous timing shortcut. We therefore
repeat our synthetic-input control in this setting. Unlike word decoding,
synthetic inputs perform only slightly above chance, reaching 6.5\% balanced
character accuracy on validation and 6.8\% on test, compared with 37.0\% and
34.9\% using real MEG. Thus, while aligned typing windows may contain some
timing information, it cannot explain the strong performance of Brain2Qwerty.

\begin{figure}
    \centering
    \includegraphics[width=0.65\linewidth]{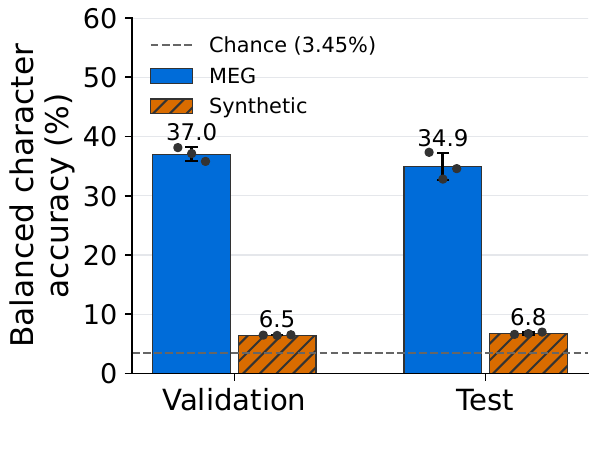}
    \caption{\textbf{The timing shortcut does not explain Brain2Qwerty performance.}
We repeat our synthetic-input control for Brain2Qwerty~\citep{Lvy2026NoninvasiveDO}.
Replacing MEG with synthetic signals that preserve the aligned-window structure
reduces balanced character accuracy from 37.0\% to 6.5\% on validation and from
34.9\% to 6.8\% on test, close to the 3.45\% chance level.
Dots show individual training seeds and error bars show standard deviations.}
    \label{fig:chars}
\end{figure}

\section{Data and Experimental Details}

\subsection{Data Splits}
\label{app:data}

We partition subject-0 recordings into training, validation, and test
sets as detailed in Table~\ref{tab:libribrain-session-split}, assigning
all recordings sharing a session label to the same partition across
tasks. Test sessions were selected using word counts to supply at least five distinct occurrences for every word position
in the core 100-sentence clinical benchmark while limiting the
amount of data reserved from training. The additional expanded 100 sentences
use previously unassigned occurrences from the same test sessions.
Session 11 was reserved for validation, with all
remaining sessions used for training.

\begin{table*}
\centering
\small
\setlength{\tabcolsep}{5pt}
\renewcommand{\arraystretch}{1.05}
\begin{tabularx}{\textwidth}{@{}rlXrr@{}}
Session ID & Split & Tasks & Recordings & Hours \\
\midrule
1 & Train & MOCHATIMIT, Sherlock1--9, TIMIT, TheMoth & 12 & 6.96 \\
2 & Train & MOCHATIMIT, Sherlock1--9, TIMIT, TheMoth & 12 & 6.85 \\
3 & Train & MOCHATIMIT, Sherlock1--9, TIMIT, TheMoth & 12 & 6.45 \\
4 & Train & MOCHATIMIT, Sherlock1--9, TIMIT, TheMoth & 12 & 6.47 \\
5 & Train & Sherlock1--9, TIMIT, TheMoth & 11 & 5.51 \\
6 & Train & Sherlock1--9, TIMIT, TheMoth & 11 & 6.61 \\
8 & Train & Sherlock1--9, TIMIT, TheMoth & 11 & 6.39 \\
9 & Train & Sherlock1--9, TIMIT, TheMoth & 11 & 7.02 \\
10 & Train & Sherlock1--9, TIMIT, TheMoth & 11 & 5.77 \\
12 & Train & Sherlock1--7, Sherlock9, TIMIT, TheMoth & 10 & 6.43 \\
17 & Train & TheMoth & 1 & 0.25 \\
19 & Train & TheMoth & 1 & 0.18 \\
23 & Train & TheMoth & 1 & 0.24 \\
24 & Train & TheMoth & 1 & 0.24 \\
25 & Train & TheMoth & 1 & 0.29 \\
26 & Train & TheMoth & 1 & 0.24 \\
27 & Train & TheMoth & 1 & 0.13 \\
28 & Train & TheMoth & 1 & 0.24 \\
\midrule
11 & Validation & Sherlock1--7, Sherlock9, TIMIT, TheMoth & 10 & 5.57 \\
\midrule
0 & Test & Sherlock9 & 1 & 0.08 \\
7 & Test & Sherlock1--9, TIMIT, TheMoth & 11 & 7.00 \\
13 & Test & Sherlock5--7, TIMIT, TheMoth & 5 & 2.63 \\
14 & Test & Sherlock5--7, TIMIT, TheMoth & 5 & 2.84 \\
15 & Test & Sherlock5, TheMoth & 2 & 0.70 \\
16 & Test & TheMoth & 1 & 0.22 \\
18 & Test & TheMoth & 1 & 0.18 \\
20 & Test & TheMoth & 1 & 0.23 \\
21 & Test & TheMoth & 1 & 0.21 \\
22 & Test & TheMoth & 1 & 0.21 \\
29 & Test & TheMoth & 1 & 0.22 \\
30 & Test & TheMoth & 1 & 0.19 \\
\midrule
\multicolumn{3}{@{}l}{Total: 31 session labels} & 162 & 86.57 \\
\end{tabularx}
\caption{\textbf{LibriBrain100 subject-0 session split.}
Session IDs and task names are taken from recording filenames.
Task ranges are inclusive: for example, Sherlock1--9 denotes the nine tasks
Sherlock1 through Sherlock9. Each listed task contributes one recording for
that session ID. All recordings sharing a session label are assigned to the
same split, including across tasks. The split contains 121 training recordings
(66.29 hours), 10 validation recordings (5.57 hours), and 31 test recordings
(14.71 hours). Durations refer to complete recordings before windowing. Core and Expanded use the test partition;
development sentences use the validation partition.}
\label{tab:libribrain-session-split}
\end{table*}

\subsection{Sentence Decoding Control Details}
\label{app:expdetails}

\paragraph{LM-only baseline.}
To measure how much of the benchmark can be solved from linguistic priors alone,
we run the same sentence-decoding procedure with the neural evidence removed.
The baseline uses the frozen Qwen3-8B-Base model with the same clinical prompt as the corresponding Brain + LM
condition. Beam search generates each candidate prefix autoregressively, with
no access to MEG signals or reference words. The only difference from Brain + LM is therefore the absence of the neural
score. Because decoding is deterministic once the prompt and sentence length
are fixed, we report a single LM-only result. This baseline quantifies how much performance is
explained by the language model under the vocabulary and length constraints of
our benchmark.

\paragraph{Noise control.}
The language model may generate plausible clinical sentences even when the
neural input contains no useful information. We therefore replace each MEG
window with Gaussian noise matched to the first two moments of the
preprocessed training data. We sample $M=10{,}000$ training windows without
replacement and estimate the mean and population variance separately for each
sensor $c$ and onset-relative time point $t$:
\begin{equation}
    \mu_{c,t}
    =
    \frac{1}{M}\sum_{j=1}^{M} x_{j,c,t},
    \qquad
    \sigma^2_{c,t}
    =
    \frac{1}{M}\sum_{j=1}^{M}
    \left(x_{j,c,t}-\mu_{c,t}\right)^2.
\end{equation}
Each test occurrence is then replaced by
\begin{equation}
    \tilde{x}_{c,t}
    =
    \mu_{c,t} + \sigma_{c,t}\epsilon_{c,t},
    \qquad
    \epsilon_{c,t}\sim\mathcal{N}(0,1),
\end{equation}
with independent draws across sensors, time points, and occurrences. The
resulting inputs therefore match the sensor- and time-specific mean and
variance of real MEG in expectation, but contain no information about the
perceived word.

\paragraph{Shuffled control.}
The Gaussian-noise control also removes the spatial and temporal structure of
real MEG. We therefore use a second control that preserves the neural
representations produced from real brain recordings while removing their
correspondence to the correct words. We apply a single random permutation,
chosen independently of the labels and without replacement, to the predicted
embeddings before sentence decoding. This preserves the distribution and
structure of the neural representations, but assigns them to the wrong test
occurrences. A substantial advantage for correctly aligned embeddings over
this shuffled control therefore shows that decoding depends on
example-specific evidence decoded from the brain.

\subsection{Training and Decoding Details}
\label{app:details}

Table~\ref{tab:simpleb2t-hyperparameters} summarises the preprocessing, architecture, training, and sentence decoding settings.

\begin{table}
\centering
\small
\begin{tabularx}{\linewidth}{@{}lX@{}}
Hyperparameter & Setting \\
\midrule
\multicolumn{2}{@{}l}{\textbf{Input}} \\
MEG channels / sampling rate & 306 / 50 Hz \\
Word window / baseline correction & 3 s from onset / subtract channel-wise mean of first 0.5 s \\
Bandpass filter / clipping & 0.1--40 Hz / $[-5,5]$ after scaling \\
Channel scaling & Recording-wise, channel-wise RobustScaler \citep{pedregosa2011scikit} \\
\midrule
\multicolumn{2}{@{}l}{\textbf{Model}} \\
CNN & Temporal CNN with spatial channel merging
and temporal attention pooling \citep{Defossez2022DecodingSP,dAscoli2025TowardsDI} \\
CNN depth / hidden channels & 5 / 160 \\
Kernel size / dilation period & 3 / 5 \\
Input dropout / batch normalization & 0.1 / enabled \\
Initial linear width & 512 \\
Spatial merger positional dimension & 2048 \\
Residual MLP blocks & 4; LayerNorm--Linear--GELU--Linear \\
MLP dimensions & $1024 \rightarrow 2048 \rightarrow 1024$ \\
Embedding normalisation & Unit $\ell_2$ norm before and after MLP \\
Word targets & Mean embeddings from layer 12 of T5-large \citep{raffel2020} \\
\midrule
\multicolumn{2}{@{}l}{\textbf{Training and model selection}} \\
Training observations & Individual word occurrences ($k=1$) \\
Objective & D-SigLIP \citep{zhai2023siglip, dAscoli2025TowardsDI} \\
Optimizer / learning rate & AdamW \citep{loshchilov2019} / $10^{-4}$ \\
Weight decay / batch size & 0 / 128 \\
Schedule / maximum epochs & Cosine annealing \citep{loshchilov2017cosine} / 50 \\
Early-stopping patience & 10 epochs without improvement \\
Checkpoint selection & Validation balanced word accuracy ($k=5$) \\
Training seeds & 0, 100, 200, 300, 400 \\
\midrule
\multicolumn{2}{@{}l}{\textbf{Sentence decoding}} \\
Temperature $T$ & $0.049$ \\
Individual-evidence $\alpha$ & 2 \\
Language model & Qwen3-8B-Base \citep{yang2025qwen3} \\
LM weight $\lambda$ & Default: 0.5; observation-dependent settings below \\
Beam width & 50 \\
\end{tabularx}
\caption{\textbf{SimpleB2T hyperparameters.} The ablations use $\lambda=0.5$, while other results use
$\lambda=(1.5,1,0.5,0.5,0.5)$ for $k=1,\ldots,5$.}
\label{tab:simpleb2t-hyperparameters}
\end{table}

\paragraph{Language-model scoring.}
We use the fixed prompt
\texttt{A patient communicates a short request to hospital staff.},
followed by a newline and \texttt{Patient:}. We do not apply a chat
template or insert any additional special tokens. Candidate words are
formatted with a leading space, the first word of the sentence is capitalised,
and the pronoun ``I'' is always uppercase.

If a candidate word $w$ is tokenised as $(u_1,\ldots,u_m)$, we score it by
summing the autoregressive log probabilities of its constituent tokens:
\begin{equation}
    \ell_{\mathrm{LM}}(w \mid h)
    =
    \sum_{j=1}^{m}
    \log p_{\mathrm{LM}}(u_j \mid h,u_{<j}),
\end{equation}
where $h$ contains the fixed prompt together with the current candidate
sentence prefix. Each token probability is computed using the model's full
tokenizer vocabulary.

\paragraph{Beam search.}
At each sentence position, beam search expands every selected prefix with
all 92 candidate words. For each expansion, we add the neural decoder score
and the language-model score weighted by $\lambda$. The number of word
positions is known in advance, and the decoder therefore generates exactly
one word at each position.

After the final word, we also score the end of the sentence. For each remaining
candidate sentence, we add
\begin{equation}
    \lambda \log\!\left[
        p_{\mathrm{LM}}(\texttt{.}\mid h)
        +
        p_{\mathrm{LM}}(\texttt{?}\mid h)
    \right],
\end{equation}
where both punctuation marks are single tokens and $h$ now contains the
complete candidate sentence. The highest-scoring completed candidate is
returned.

\paragraph{Language-model weight across observations.}
\label{app:weight}

\begin{figure}
    \centering
    \includegraphics[width=1.0\linewidth]{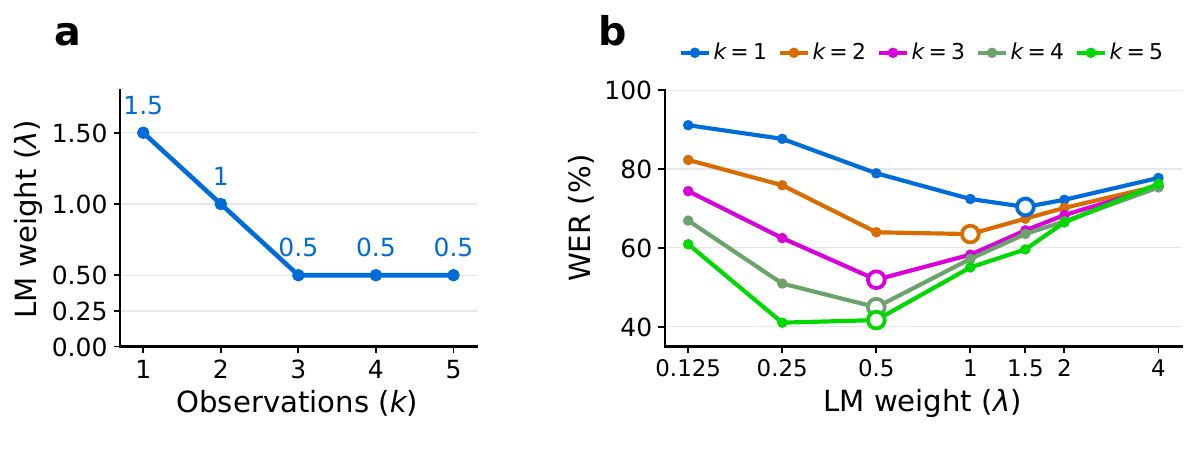}
    \caption{\textbf{Language-model weight decreases as neural evidence improves.}
\textbf{(a)} Development-selected $\lambda$ falls as more observations are aggregated.
\textbf{(b)} WER is minimised at lower LM weights for larger $k$.}
    \label{fig:lmweight}
\end{figure}

We select the language-model weight $\lambda$ separately for each number of
observations $k$ using the 50-sentence development set, and fix the selected
value before test evaluation. The selected weights are
\[
\begin{array}{c|ccccc}
k & 1 & 2 & 3 & 4 & 5 \\
\hline
\lambda & 1.5 & 1.0 & 0.5 & 0.5 & 0.5
\end{array}
\]
As more observations are aggregated, the selected language-model weight
decreases and then stabilises (Figure~\ref{fig:lmweight}).

\section{Additional SimpleB2T Analyses}

\subsection{Predictability from Subsequent Words}
\label{sec:future-predictability}

This analysis tests whether SimpleB2T benefits from information about later
words contained within each individual three-second window.

For each core test occurrence, we identify all later words whose annotated durations fall entirely within the same three-second window. We then use frozen
Qwen3-8B-Base to measure how strongly this continuation predicts the target
word. Specifically, for each of the 92 candidate target words, we compute the
likelihood of the observed continuation conditioned on that candidate, and
normalise these likelihoods assuming a uniform prior over candidates. This
gives a probability distribution over possible target words using only the
subsequent transcript. For each sentence position, we average these
distributions across its five assigned occurrences.

We next test whether SimpleB2T performs better when this subsequent context is
more predictive. Predictability varies substantially across positions, from 0.04\% to 84.6\%,
with a median of 20.7\% and an interquartile range of 8.1--36.9\%.
Thus, the benchmark contains both continuations that provide little information
about the target word and continuations that strongly constrain its identity. We split examples into low- and
high-predictability groups. Because some words may be intrinsically easier to
decode than others, we also construct a word-matched split. For each target
word separately, we rank its positions by subsequent-context predictability
and assign equal numbers to the low- and high-predictability groups.

Table~\ref{tab:future-predictability} reports positional word accuracy using
the unchanged $k=5$, beam-50 predictions. In the word-matched analysis, Brain
accuracy rises modestly from 24.0\% on low-predictability positions to 27.1\%
on high-predictability positions. In contrast, Brain + LM accuracy is almost
unchanged, at 65.3\% and 64.9\%, respectively. The overall split shows no
consistent advantage for high-predictability positions. Therefore, SimpleB2T does not appear to significantly benefit from other words in each three-second window.

\begin{table}[h]
\centering
\small
\setlength{\tabcolsep}{5pt}
\begin{tabular}{@{}lcccc@{}}
 & \multicolumn{2}{c}{Overall split}
 & \multicolumn{2}{c}{Word-matched split} \\
\cmidrule(lr){2-3}\cmidrule(lr){4-5}
Method & Low & High & Low & High \\
\midrule
LM
 & $26.3$ & $25.8$
 & $28.3$ & $28.3$ \\
Brain
 & $27.5_{\pm2.0}$ & $23.8_{\pm2.7}$
 & $24.0_{\pm2.1}$ & $27.1_{\pm2.6}$ \\
Brain + LM
 & $62.6_{\pm1.3}$ & $64.0_{\pm2.2}$
 & $65.3_{\pm2.0}$ & $64.9_{\pm1.9}$ \\
\end{tabular}
\caption{\textbf{Word accuracy grouped by the predictability of subsequent
context.} Subscripts denote sample standard
deviations across five training seeds. The LM is deterministic.}
\label{tab:future-predictability}
\end{table}

\subsection{Train--Test Mismatch Control in the Clinical Benchmark}
\label{app:traintestbench}

One possible explanation for the poor transfer of jointly decoding words to
our clinical benchmark is a mismatch between training and evaluation. The
original model is trained on sentences with overlapping neighbouring
windows, whereas benchmark positions are assembled from separately recorded
occurrences and therefore do not overlap. To test this, we retrain the same
model on non-overlapping sentence constructions matched to the
evaluation setting.

\looseness=-1 We construct non-overlapping versions of the natural training
sentences while preserving their original word sequences. For each word
position, we replace the original MEG window with a three-second window from a
different recorded occurrence of the same word. We require that no two windows
assigned to the same constructed sentence overlap in the underlying recording.
They must either come from different recordings or begin at least three seconds
apart. The model therefore sees exactly the same linguistic sequence as before,
but cannot recover word intervals from shared MEG samples.

To construct these sentences without reusing data, we permute recorded
occurrences among positions of the same word type, so that each retained
occurrence is assigned to exactly one position per epoch. We then resolve any
remaining overlaps within sentences by swapping assignments between positions
with the same target word. Sentences for which no valid assignment can be found
are discarded. This retains \emph{96.3\% of eligible training occurrences}.
Assignments are re-randomised between epochs while preserving the same
non-overlap constraint.

Retraining does not materially improve sentence reconstruction (Table~\ref{tab:stitched-clinical}). With LLM
rescoring at $k=5$, the matched model obtains 79.2\%, 85.5\%, and 82.6\% WER
on Core, Expanded, and Full, respectively, compared with 79.4\%, 85.5\%, and
82.7\% for the original overlap-trained model. Our clinical benchmark is
constructed from separately recorded occurrences at each sentence position, so
there is no coherent cross-word neural activity for the \citeauthor{dAscoli2025TowardsDI} decoder to
exploit. The control shows that matching the decoder to
the non-overlapping, word-by-word setting does not recover the performance of
independent decoding.

\begin{table*}[h]
\centering
\footnotesize
\setlength{\tabcolsep}{3pt}
\renewcommand{\arraystretch}{1.05}
\begin{tabular*}{\textwidth}{@{\extracolsep{\fill}}lc*{6}{c}@{}}
 & & \multicolumn{2}{c}{Core} & \multicolumn{2}{c}{Expanded} & \multicolumn{2}{c}{Full} \\
\cmidrule(lr){3-4}\cmidrule(lr){5-6}\cmidrule(lr){7-8}
Method & $k$ & WER (\%) $\downarrow$ & \shortstack{SMR (\%) $\uparrow$} & WER (\%) $\downarrow$ & \shortstack{SMR (\%) $\uparrow$} & WER (\%) $\downarrow$ & \shortstack{SMR (\%) $\uparrow$} \\
\midrule
d'Ascoli & 1 & $97.7_{\pm0.4}$ & $0.0_{\pm0.0}$ & $97.0_{\pm0.7}$ & $0.0_{\pm0.0}$ & $97.3_{\pm0.6}$ & $0.0_{\pm0.0}$ \\
d'Ascoli + LM & 1 & $79.9_{\pm1.5}$ & $2.3_{\pm0.2}$ & $86.2_{\pm0.9}$ & $0.6_{\pm0.0}$ & $83.2_{\pm0.2}$ & $1.4_{\pm0.1}$ \\
\midrule
d'Ascoli & 5 & $98.0_{\pm0.6}$ & $0.0_{\pm0.0}$ & $97.2_{\pm1.1}$ & $0.0_{\pm0.0}$ & $97.5_{\pm0.8}$ & $0.0_{\pm0.0}$ \\
d'Ascoli + LM & 5 & $79.2_{\pm1.6}$ & $2.3_{\pm0.6}$ & $85.5_{\pm0.7}$ & $1.0_{\pm0.0}$ & $82.6_{\pm0.5}$ & $1.7_{\pm0.3}$ \\
\end{tabular*}
\caption{\textbf{Controlling for train--test mismatch in the joint word decoder.}
We retrain the decoder of \citet{dAscoli2025TowardsDI} on non-overlapping
sentence constructions and evaluate at
$k=5$. Jointly decoding words remains much less
accurate than SimpleB2T. Values are means across training seeds and subscripts
show sample standard deviations.}
\label{tab:stitched-clinical}
\end{table*}

\subsection{Word-level Analysis}
\label{app:word-analysis}

\begin{figure}
    \centering
    \includegraphics[width=1.0\linewidth]{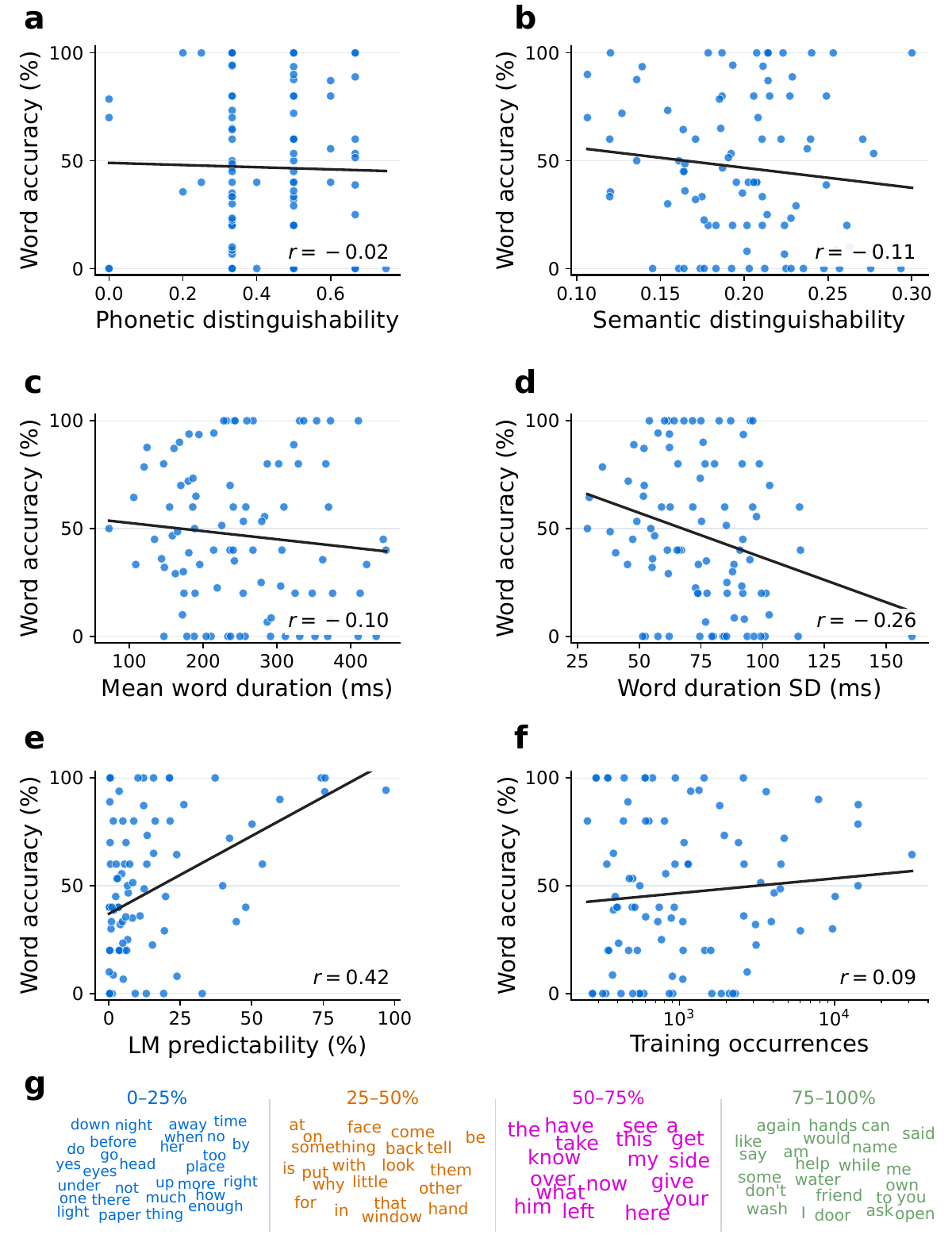}
\caption{\textbf{Word-level decoding analysis.}
Word accuracy is weakly related to \textbf{(a)} phonetic distinguishability,
\textbf{(b)} semantic distinguishability, and \textbf{(c)} mean word duration.
It is negatively related to \textbf{(d)} word duration variability and
positively related to \textbf{(e)} LM predictability, with little association
with \textbf{(f)} training frequency.
\textbf{(g)} Words grouped by accuracy bins.}
    \label{fig:word-analysis}
\end{figure}

We further examine which words are easiest to decode.
Figure~\ref{fig:word-analysis} relates mean Brain + LM word accuracy
across five training seeds to properties of the 92 vocabulary words.
Phonetic distinguishability is the minimum phoneme edit distance to
another vocabulary word, normalised by the longer pronunciation's length,
using CMUdict pronunciations with stress markers removed and taking the
minimum if there are pronunciation variants.
Semantic distinguishability is one minus the maximum cosine similarity
to another word's frozen T5 target embedding.
Neither measure is strongly associated with accuracy
($r=-0.02$ and $r=-0.11$, respectively).
Training frequency ($r=0.09$) and mean word duration ($r=-0.10$) also
show weak associations.
Duration variability, computed as the sample standard deviation of
annotated word durations across the used training occurrences, has a
stronger negative association ($r=-0.26$). Therefore, words with more consistent
durations tend to be decoded more accurately. The strongest association among these measures is with LM predictability ($r=0.42$), defined as the target word's probability under the frozen LM, conditioned on the fixed prompt and preceding reference words. All correlations are Pearson correlations. The accuracy-binned word groups illustrate the range of performance, with both function
and content words appearing across accuracy bins.

\subsection{Effect of the Linguistic Prior}
\label{app:prior}

We examine how the language-model prior changes with the prompt (Table~\ref{tab:prompt-ablation}).
These experiments were conducted \emph{after} all main SimpleB2T experiments
were complete and were not used to select the prompt reported in the main
results. The prompt used throughout the paper was fixed before this ablation. We compare our prompt with several alternatives ranging from a minimal prefix to broader descriptions of English, everyday speech,
and clinical communication. The results show that specifying the intended
patient-communication setting improves decoding relative to more
generic prompts. This supports the view that part of SimpleB2T's success comes
from using a linguistic prior matched to the constrained communication task. Future work may benefit from optimising the prompt on a development set to further improve the linguistic prior.

\begin{table*}[h]
\centering
\footnotesize
\setlength{\tabcolsep}{3pt}
\renewcommand{\arraystretch}{1.12}
\begin{tabular*}{\textwidth}{@{\extracolsep{\fill}}lc*{6}{c}@{}}
& & \multicolumn{2}{c}{Core} & \multicolumn{2}{c}{Expanded} & \multicolumn{2}{c}{Full} \\
\cmidrule(lr){3-4}\cmidrule(lr){5-6}\cmidrule(lr){7-8}
Prompt & $\lambda$ & LM only & Brain + LM & LM only & Brain + LM & LM only & Brain + LM \\
\midrule
A & 0.25 & $97.3$ & $55.2_{\pm1.2}$ & $98.1$ & $53.9_{\pm0.8}$ & $97.7$ & $54.5_{\pm1.0}$ \\
B & 0.5 & $96.0$ & $51.3_{\pm3.6}$ & $91.0$ & $52.5_{\pm2.6}$ & $93.3$ & $51.9_{\pm3.1}$ \\
C & 0.25 & $91.8$ & $51.8_{\pm3.3}$ & $90.0$ & $49.8_{\pm3.1}$ & $90.8$ & $50.7_{\pm3.2}$ \\
D & 0.25 & $84.4$ & $49.3_{\pm2.0}$ & $95.0$ & $48.1_{\pm3.2}$ & $90.0$ & $48.7_{\pm2.6}$ \\
E (ours) & 0.5 & \cellcolor{gray!15}$\mathbf{73.8}$ & \cellcolor{gray!15}$\mathbf{36.6_{\pm1.7}}$ & $91.8$ & $44.8_{\pm3.7}$ & $83.4$ & $41.0_{\pm2.2}$ \\
F & 0.5 & $78.8$ & $38.7_{\pm4.3}$ & \cellcolor{gray!15}$\mathbf{87.2}$ & \cellcolor{gray!15}$\mathbf{42.6_{\pm5.1}}$ & \cellcolor{gray!15}$\mathbf{83.3}$ & \cellcolor{gray!15}$\mathbf{40.8_{\pm4.6}}$ \\
\end{tabular*}
\par\vspace{0.8em}
\begin{tabularx}{\textwidth}{@{}l>{\raggedright\arraybackslash}X@{}}
ID & Exact prompt \\
\midrule
A & Sentence: \\
B & The following is a sentence from an English passage: \\
C & The following is a sentence from everyday speech: \\
D & The following is a sentence spoken in a clinical setting: \\
E & A patient communicates a short request to hospital staff.\newline Patient: \\
F & A patient who cannot speak communicates a short message to hospital staff about comfort, positioning, personal care, their surroundings, or interaction with other people.\newline Patient: \\
\end{tabularx}
\caption{\textbf{Prompt specificity improves sentence reconstruction.}
Brain + LM uses $k=5$, beam width 50, and frozen Qwen3-8B-Base. Subscripts report sample standard deviations. LM only decoding is deterministic. Bold, shaded cells mark the lowest WER within each condition and split. A line break before \texttt{Patient:} is part of prompts E and F.}
\label{tab:prompt-ablation}
\end{table*}

\subsection{OVMI Scores}

\looseness=-1 WER depends on the vocabulary and language distribution used for evaluation,
making it difficult to compare performance across communication settings and with other work.
We therefore also report OVMI \citep{jayalath2026common}, which measures the
information conveyed by a decoder relative to a reference communication
distribution. Table~\ref{tab:normalized-ovmi} reports normalised OVMI under
four reference distributions.

\begin{table}[h]
\centering
\small
\setlength{\tabcolsep}{5pt}
\begin{tabular}{lc ccc}
Reference $p$ & $k$ & Core & Expanded & Full \\
\midrule
\multirow{2}{*}{SUBTLEX-UK}
 & 1 & $1.40_{\pm0.09}$ & $1.05_{\pm0.03}$ & $1.23_{\pm0.05}$ \\
 & 5 & $6.41_{\pm0.48}$ & $6.09_{\pm0.83}$ & $6.25_{\pm0.54}$ \\
\addlinespace
\multirow{2}{*}{Switchboard}
 & 1 & $1.88_{\pm0.12}$ & $1.41_{\pm0.04}$ & $1.66_{\pm0.06}$ \\
 & 5 & $8.53_{\pm0.63}$ & $8.11_{\pm1.09}$ & $8.32_{\pm0.71}$ \\
\addlinespace
\multirow{2}{*}{UCV}
 & 1 & $5.68_{\pm0.36}$ & $4.29_{\pm0.13}$ & $5.04_{\pm0.18}$ \\
 & 5 & $24.32_{\pm1.71}$ & $23.17_{\pm2.97}$ & $23.72_{\pm1.93}$ \\
\addlinespace
\multirow{2}{*}{Sherlock}
 & 1 & $1.24_{\pm0.08}$ & $0.93_{\pm0.03}$ & $1.10_{\pm0.04}$ \\
 & 5 & $5.65_{\pm0.42}$ & $5.37_{\pm0.73}$ & $5.50_{\pm0.47}$ \\
\end{tabular}
\caption{\textbf{SimpleB2T normalised OVMI scores across reference distributions.} OVMI measures the information conveyed per word by a model relative to a reference communication distribution.
Values are $100\times\mathrm{OVMI}/H(p)$, where $H(p)$ is the entropy
of the full reference distribution. These estimates use balanced word accuracies of
16.04/13.29/14.80\% at $k=1$ and 46.83/45.08/45.93\% at $k=5$
for Core/Expanded/Full, respectively.
Results report means across five training seeds, with sample
standard deviations as subscripts. See \citet{jayalath2026common} for further details on OVMI.}
\label{tab:normalized-ovmi}
\end{table}

\section{Clinical Communication Benchmark}
\label{app:benchmark}

Tables~\ref{tab:core-sentences} (core) and~\ref{tab:expanded-sentences} (expanded) list the
200 clinically motivated sentences used in our communication benchmark. The sentences were generated with assistance from
GPT-6 Astra \citep{openai2026astra} and subsequently reviewed to ensure they represented plausible
patient communication needs. The MEG samples for the benchmark were constructed by assigning recorded
word occurrences to positions in predefined sentences. For each
position, we select five distinct occurrences of the required word from
the held-out test sessions, irrespective of their original sentence
context, and extract the corresponding word-aligned MEG windows. Each
occurrence is assigned to exactly one position across the 200-sentence
benchmark, so no recording event is reused.

\begin{longtable}{@{}r >{\raggedright\arraybackslash}p{0.39\linewidth}
                     r >{\raggedright\arraybackslash}p{0.39\linewidth}@{}}
\caption{\textbf{Core communication benchmark sentences.}}
\label{tab:core-sentences}\\
ID & Sentence & ID & Sentence \\
\midrule
\endfirsthead

\multicolumn{4}{c}{\tablename~\thetable\ continued from previous page} \\
\toprule
ID & Sentence & ID & Sentence \\
\midrule
\endhead

\midrule
\multicolumn{4}{r}{Continued on next page} \\
\endfoot

\endlastfoot

1  & Can you help me? &
51 & Can you open the door? \\

2  & I would like some help. &
52 & Can you open the window? \\

3  & Can you come here? &
53 & Can you put the light on? \\

4  & Can you come back? &
54 & Can you take this away? \\

5  & Can you be here with me? &
55 & Can you put that over me? \\

6  & Can you give me more time? &
56 & Can you put this by my hand? \\

7  & I would like to be on my own. &
57 & Can you put this on the other side? \\

8  & Do not go away. &
58 & Can you put that down? \\

9  & Can you come back in a little while? &
59 & Can you put that here? \\

10 & Can you ask for help? &
60 & Can you put that back? \\

11 & I would like some water. &
61 & Can you tell me your name? \\

12 & Can you give me some water? &
62 & Can you tell me what that is? \\

13 & I would like more water. &
63 & Can you tell me what this is for? \\

14 & That is enough water. &
64 & Can you say that again? \\

15 & Can you put the water here? &
65 & Can you tell me more? \\

16 & Can you help me with the water? &
66 & Can you give me some paper? \\

17 & Can you take the water away? &
67 & Can you look at this paper? \\

18 & I would like a little more. &
68 & I don't see that. \\

19 & That is too much. &
69 & I don't see your face. \\

20 & No more for now. &
70 & Can you ask me one thing at a time? \\

21 & Can you help me up? &
71 & Can you give me a little time? \\

22 & Can you help me down? &
72 & I have something to say. \\

23 & Can you put my head up? &
73 & That is not what I said. \\

24 & Can you put my head down? &
74 & Yes, that is right. \\

25 & Can you put my hand here? &
75 & No, that is not right. \\

26 & Can you put my hand there? &
76 & I do not know. \\

27 & Can you take my hand? &
77 & I would like to know why. \\

28 & Can you put this under my head? &
78 & Can you tell me when? \\

29 & Can you put this under my back? &
79 & Can you tell me how? \\

30 & Can you help me get on my side? &
80 & I would like to ask something. \\

31 & I would like to be on my left side. &
81 & I would like to see my friend. \\

32 & I would like to be on my right side. &
82 & Can you tell my friend to come here? \\

33 & I would like to be on my back. &
83 & Can you ask my friend to come back? \\

34 & Can you help me with my head? &
84 & Can you tell him I am here? \\

35 & Can you help me with my back? &
85 & Can you tell her I am here? \\

36 & Can you help me with my left hand? &
86 & I would like to see him. \\

37 & Can you help me with my right hand? &
87 & I would like to see her. \\

38 & Do not put that on my head. &
88 & Can you ask them to come in? \\

39 & Do not put that on my back. &
89 & Can you ask them to come back? \\

40 & That is not the right place. &
90 & I would like more time with them. \\

41 & Can you wash my face? &
91 & Can you help me now? \\

42 & Can you wash my hands? &
92 & That is too much light. \\

43 & Can you wash my head? &
93 & There is something in my eyes. \\

44 & Can you wash my back? &
94 & There is something on my face. \\

45 & Can you wash my left hand? &
95 & Can you look at my hand? \\

46 & Can you wash my right hand? &
96 & Can you look at my head? \\

47 & Can you wash here? &
97 & Can you look at my back? \\

48 & Can you wash there? &
98 & Can you look at my eyes? \\

49 & Can you help me wash? &
99 & Can you tell me the time? \\

50 & Do not get water in my eyes. &
100 & Can you come back before night? \\

\end{longtable}

\begin{longtable}{@{}r >{\raggedright\arraybackslash}p{0.39\linewidth}
                     r >{\raggedright\arraybackslash}p{0.39\linewidth}@{}}
\caption{\textbf{Expanded communication benchmark sentences.}}
\label{tab:expanded-sentences}\\
ID & Sentence & ID & Sentence \\
\midrule
\endfirsthead

\multicolumn{4}{c}{\tablename~\thetable\ continued from previous page} \\
\toprule
ID & Sentence & ID & Sentence \\
\midrule
\endhead

\midrule
\multicolumn{4}{r}{Continued on next page} \\
\endfoot

\endlastfoot

1  & I would like some water now. &
51 & Tell her to come back in a little while. \\

2  & Give me a little water now. &
52 & Tell him to come back before night. \\

3  & I would like water before you go. &
53 & I would like to see them now. \\

4  & Take the water away now. &
54 & I would like to have more time with her. \\

5  & I have enough water for now. &
55 & I would like to have more time with him. \\

6  & I would like the water by my hand. &
56 & Ask them to be here with me. \\

7  & Do not give me more water. &
57 & I do not know your name. \\

8  & Give me some more time with the water. &
58 & Tell me your name again. \\

9  & There is water on my face. &
59 & Tell me why you have come back. \\

10 & There is water in my eyes. &
60 & I would like to know what this is. \\

11 & I would like to wash my face. &
61 & I would like to know what that is for. \\

12 & I would like to wash my hands. &
62 & I do not know why you said that. \\

13 & Wash my face with a little water. &
63 & Say that one more time. \\

14 & Do not wash my eyes. &
64 & Say one thing at a time. \\

15 & Wash my hands again. &
65 & Give me time to say what I would like. \\

16 & I would like to wash before night. &
66 & I have more to say. \\

17 & Wash the other hand too. &
67 & That is what I would like. \\

18 & There is something under my back. &
68 & That is not what I would like. \\

19 & There is something under my head. &
69 & I said no. \\

20 & I would like my head a little more up. &
70 & I said yes. \\

21 & I would like my head a little more down. &
71 & I would like to ask you something. \\

22 & My hand is not in the right place. &
72 & Ask me again in a little while. \\

23 & Take this away now. &
73 & Do not ask me now. \\

24 & I would like to be on the other side. &
74 & I do not know how to say this. \\

25 & Do not take my hand away. &
75 & I would like some paper now. \\

26 & Put this under my hand. &
76 & Give me the paper again. \\

27 & Put that by my side. &
77 & Look at what is on the paper. \\

28 & Put the water on the other side. &
78 & I don't see the paper. \\

29 & Put this paper by my hand. &
79 & I would like the paper here. \\

30 & I would like my back down a little. &
80 & There is something I would like you to see. \\

31 & The light is in my eyes. &
81 & Look at my face when I say this. \\

32 & I would like a little more light. &
82 & I don't see what is in your hand. \\

33 & I would like the light on now. &
83 & I would like to see your face. \\

34 & Do not have the light on at night. &
84 & I don't know what to do. \\

35 & The window is open. &
85 & Tell me what you would like me to do. \\

36 & I would like the window open for a little while. &
86 & Tell me before you wash my face. \\

37 & Do not open the window now. &
87 & Tell me before you take this away. \\

38 & I would like the door open. &
88 & I would like to know when you would be back. \\

39 & Do not open the door now. &
89 & How much time do I have? \\

40 & Do not have this by my eyes. &
90 & What is this water for? \\

41 & Come here for a little while. &
91 & Is that my name on the paper? \\

42 & Be here while I have some water. &
92 & Is my friend here now? \\

43 & I would like you here with me. &
93 & Do you know when my friend would be here? \\

44 & Do not go while I have something to say. &
94 & Do you know when I would see him again? \\

45 & Come back at night. &
95 & I would like you to look at my eyes. \\

46 & Come back in a little while. &
96 & I would like you to look at my back. \\

47 & I would like a little time on my own. &
97 & Do not give me that now. \\

48 & I would like my friend here with me. &
98 & That is enough for now. \\

49 & Ask my friend to come in. &
99 & I would like more time before you go. \\

50 & Ask my friend to come back at night. &
100 & I would like to know why you said no. \\

\end{longtable}

\end{document}